%% file: article.tex
\RequirePackage{amsthm}

\documentclass[sn-mathphys-num]{sn-jnl}

\usepackage[english]{babel}
\usepackage{graphicx}%
\usepackage{multirow}%
\usepackage{amsmath,amssymb,amsfonts}%
\usepackage{amsthm}%
\usepackage{bm}
\usepackage{mathrsfs}%
\usepackage[title]{appendix}%
\usepackage[table]{xcolor}%
\usepackage{textcomp}%
\usepackage{manyfoot}%
\usepackage{booktabs}%
\usepackage{listings}%
\usepackage{subcaption}
\usepackage{lipsum}
\usepackage{adjustbox}
\usepackage{rotating,tabularx,siunitx}
\usepackage{mathtools}
\usepackage{cleveref}
\usepackage{algorithm2e}
\usepackage[siunitx]{circuitikz}
\usepackage{pdfpages}
\usepackage{comment}
\usepackage{lmodern}
\usepackage{textcomp}

\usepackage{tikz}
\usepackage{bm}
\usetikzlibrary{arrows.meta}
\usetikzlibrary{math} 
\usetikzlibrary{calc}

\RestyleAlgo{ruled}
\SetKwComment{Comment}{/* }{ */}

\usepackage{colortbl}
\usepackage{pgfplots}
\pgfplotsset{compat=newest}

\usepackage{longtable}

\theoremstyle{thmstyleone}%
\theoremstyle{thmstyletwo}%

\theoremstyle{thmstylethree}%

\begin{document}

\title[Article Title]{Time series classification with random convolution kernels: pooling operators and input representations matter}


\author*[1]{\fnm{Mouhamadou Mansour} \sur{Lo}}\email{mouhamadou.lo@univ-artois.fr}

\author[2]{\fnm{Gildas} \sur{Morvan}}\email{gildas.morvan@univ-artois.fr}

\author[1]{\fnm{Mathieu} \sur{Rossi}}\email{mathieu.rossi@univ-artois.fr}

\author[1]{\fnm{Fabrice} \sur{Morganti}}\email{fabrice.morganti@univ-artois.fr}

\author[2]{\fnm{David} \sur{Mercier}}\email{david.mercier@univ-artois.fr}

\affil[1]{Univ. Artois, UR 4025, Laboratoire Systèmes Electrotechniques et Environnement (LSEE), F-62400 Béthune, France}

\affil[2]{Univ. Artois, UR 3926, Laboratoire de Génie Informatique et d'Automatique de l'Artois (LGI2A), F-62400 Béthune, France}




\abstract{Time series classification (TSC) is a fundamental problem in supervised learning, with applications in domains ranging from healthcare and finance to industrial monitoring. Recent advances based on random convolution kernels, such as ROCKET, MiniRocket, MultiRocket, and HYDRA, have demonstrated that large sets of randomly generated convolutional features combined with simple linear classifiers can achieve state-of-the-art accuracy at very low computational cost. These methods typically rely on fixed input representations and a small number of pooling operators, most commonly the proportion of positive values (PPV), to extract features from activation maps. However, the optimal choice of pooling operator and representation can vary substantially across datasets. In this work, we first provide an empirical analysis on the UCR archive showing that using a fixed pooling operator does not generally yield optimal performance, and that the best choice depends on the data. Motivated by this observation, we propose SelF-Rocket (Selected Features Rocket), a novel extension of MiniRocket that dynamically attempts to select the most appropriate combination of input representation and pooling operator during training. SelF-Rocket incorporates a feature selection module that adapts to each dataset, allowing it to select the most appropriate pooling operator and representation. Extensive experiments on the UCR benchmark demonstrate that SelF-Rocket achieves competitive or superior accuracy compared with existing random convolution kernel methods.}




\maketitle

\section{Introduction}\label{Introduction}






Time Series Classification (TSC) is a type of supervised learning problem where the objective is to assign a class label to a given time series instance. The data in a time series is made up of sequences of observations collected at constant intervals over time. Formally, a time series $\mathbf{X}$ can be represented as:
\begin{equation}
   \mathbf{X} = \{x_1, x_2, \ldots, x_T\}~, 
\end{equation}
where $x_t$ denotes the value of the time series at time step $t$, and $T$ represents the length of the time series. This notation will be used to refer to a time series throughout this article.

Many methods of various kinds have been developed to address TSC problems. Traditional methods include distance-based techniques like Dynamic Time Warping (DTW)~\cite{sakoe1978dynamic}, which measure the similarity between time series. More recently, Deep Learning approaches such as Inceptiontime~\cite{ismail2020inceptiontime} have been proposed. Hybrid approaches like HIVE-COTE v2.0, combining multiple classifiers of different kinds, are generally the most accurate, at a high computing cost~\cite{middlehurst2021hive}. For a recent general survey of TSC methods, the interested reader may refer to~\cite{middlehurst_bake_2024}.

Among all these methods, Random convolution kernels based transforms, ROCKET~\cite{dempster2020rocket} and its successors, MiniRocket~\cite{dempster2021minirocket}, MultiRocket~\cite{tan2022multirocket} and HYDRA~\cite{dempster2023hydra} have attracted significant interest in recent years. The main idea behind these methods is to use a very large number of random convolution kernels to transform the time series and apply one or more pooling operators to generate a set of synthetic features that are subsequently fed into a simple linear classifier. The training phase thus only involves learning the appropriate weights assigned to each feature, which allows these methods to be, at the time this article is written, among the fastest while achieving state-of-the-art accuracy~\cite{middlehurst_bake_2024}.

However, not all kernels contribute positively to model performance, resulting in unnecessary computational complexity~\cite{salehinejad2022s}.
Thus, several methods such as {S-ROCKET}~\cite{salehinejad2022s}, {POCKET}~\cite{chen2024pocket} or {Detach-ROCKET}~\cite{uribarri2024detach} have been suggested for pruning the kernels, retaining only those that are most relevant to the problem.\

Another approach to improving ROCKET-based classifiers is through ensemble methods. Arsenal~\cite{middlehurst2021hive} combines multiple lightweight ROCKET classifiers to improve probability estimation and serves as a replacement for ROCKET within HIVE-COTE v2.0, using majority voting for classification. FT-FVC~\cite{he2023ft} enhances ROCKET by increasing feature diversity via multiple signal transformations, training separate classifiers on each representation and aggregating predictions through hard voting.

In this article, an alternative way of improving ROCKET-based classifiers is considered. Indeed, to the best of our knowledge, all the proposed transformations rely on one or more fixed pooling operators such as, for example, PPV (\textit{Proportion of positive values}) and input representations to extract features from time series.

The first contribution of this work is an experimental demonstration that this approach does not yield the best accuracy, and that adaptively selecting the most suitable pooling operator and input representation based on the data is preferable.We further provide insights into the conditions under which specific pooling operators yield superior performance.

In light of this analysis, the second contribution, a new approach based on MiniRocket, called SelF-Rocket (\textbf{Sel}ected \textbf{F}eatures Rocket), is developed. Unlike existing ones based on random convolution kernels, SelF-Rocket dynamically attempts to select the best combination of data input representations and pooling operator during the training process. The complete source code of our implementation of SelF-Rocket is available on GitHub\footnote{\url{https://github.com/ANR-MYEL/SelF-Rocket/}}.

This article is organized as follows. The main random convolution kernels based methods for TSC are first recalled in Section~\ref{sec:relatedworks}. Then, a statistical analysis, conducted on the University of California Riverside (UCR) archive~\cite{dau_ucr_2018}, is presented in Section~\ref{sec:importance} to motivate this work. It shows that while PPV is a pooling operator of choice, in most cases it is not the best one. The method SelF-Rocket is then introduced in Section~\ref{sec:selfrocket} and its performances on the UCR archive are analyzed in Section~\ref{sec:experiments}. Finally, Section~\ref{sec:conclusionandprospects} gives a conclusion with some perspectives.

\section{Main random convolution kernel–based methods for TSC}\label{sec:relatedworks}


\textbf{ROCKET} (\textit{Random Convolutional Kernel Transform})~\cite{dempster2020rocket} is the first algorithm of this kind to have been introduced. It randomly generates a large number of convolution kernels (typically 10,000) which it uses to create activation maps. These maps are then summarized by two pooling operators: PPV (\textit{Proportion of Positive Values}), which computes the percentage of positive values, and GMP (\textit{Global Maximum Pooling}), which extracts the maximum value from the activation map. Formally:
\begin{equation}
    PPV(Z) = \frac{1}{n}\sum_{i=1}^{n}[z_i>0], 
    \label{eqn:ppv}
\end{equation}
\begin{equation}
    GMP(Z) = max(Z), 
    \label{eqn:gmp}
\end{equation}
where $Z$ is the convolution output of $X$.

For each kernel, two features are extracted using PPV and GMP.

Although random, the kernels are parameterized as follows:
\begin{itemize}
    \item Length is randomly chosen from $\{7, 9, 11\}$ with equal probability.
    \item Weights $w \sim \mathcal{N}(0,1)$ are randomly selected and then normalized: $w = W - \overline{W}$.
    \item A bias $b \sim \mathcal{U}(-1,1)$ is added to the activation map.
    \item Dilations are computed with $d = \lfloor 2^x \rfloor$, where $x \sim \mathcal{U}(0,A)$ and $A = \log_2\left(\frac{\ell_{input} - 1}{\ell_{kernel} - 1}\right)$, where $\ell_{input}$ is the length of the input time series, and $\ell_{kernel}$ the length of the kernel.
\end{itemize}

\textbf{MiniRocket}~\cite{dempster2021minirocket} is a variant of ROCKET with several key modifications that greatly speed-up the training without sacrificing performances. The method becomes much more deterministic by making the following changes:
\begin{itemize}
    \item Kernel length is fixed at 9 instead of $\{7, 9, 11\}$.
    \item It uses a single set of 84 kernels containing only $\{-1, 2\}$ as values. This major change optimizes convolution operations and significantly reduces computation time.
    \item The bias is now derived from the convolution result with a kernel/dilation pair.
    \item GMP is no longer calculated, only PPV is used.
\end{itemize}

\textbf{MultiRocket}~\cite{tan2022multirocket} extends MiniRocket by introducing several improvements:
\begin{itemize}
    \item First-order difference ($DIFF$) between two time units, corresponding to the rate of change, is used as an additional input representation. Formally:
    \begin{equation}
    DIFF(X) {=}\{x_t{-}x_{t-1}: \forall t \in \{2,...,T\}\}
    \label{eqn:difference}
    \end{equation}

    \item In addition to PPV, the following features are extracted: Mean of Positive Values (MPV), Mean of Indices of Positive Values (MIPV) and Longest Stretch of Positive Values (LSPV), formally defined by:
    \begin{equation}
        MPV(Z) = \frac{1}{m}\sum_{i=1}^{m}z^{+}_i,
        \label{eqn:mpv}
    \end{equation}
     where $Z^{+} = \{ z^{+}_1, ... z^{+}_m \}$ is the subset of positive value in $Z$,
     \begin{equation}
     MIPV(Z) = 
    \begin{dcases}
    \frac{1}{m}\sum_{j=1}^{m}i^{+}_j & \text{if } m > 0 \\
    -1 & \text{otherwise}
    \end{dcases},
    \label{eqn:mipv}
    \end{equation}
     where $I^{+} = \{ i^{+}_1, ... i^{+}_m \}$ indicates the indices of positive values, and 
     \begin{equation}
    LSPV(Z) = max(j-i\mid \forall_{i\leq k\leq j}z_k>0).
    \label{eqn:lspv}
\end{equation}
\end{itemize}

\textbf{HYDRA} (\textit{Hybrid Dictionary-ROCKET Architecture})~\cite{dempster2023hydra} combines aspects of dictionary-based methods and random convolutions. The general idea is to group all kernels into groups of kernels. For each time series in the dataset, convolution is performed with all kernels from a given group. At each point in the series, the kernel leading to the highest convolution value is selected. Thus, for each group, a histogram of maximum kernel responses is created, serving as features for the classifier. First-order difference is used here as well. The kernels are parameterized as follows:
\begin{itemize}
    \item Length is fixed at 9.
    \item Weights $w \sim \mathcal{N}(0,1)$ are randomly selected.
    \item No bias or pooling operator is used after convolution.
\end{itemize}
The features generated by HYDRA and MultiRocket can be concatenated leading to a more accurate classification~\cite{dempster2023hydra}. 

\indent\\

\textbf{Alternative pooling operators} may also be applied, including Zero Crossing (ZC), which counts the number of sign changes in the activation map~\cite{sapsanis2013improving}:
\begin{equation}
\begin{aligned}
ZC(Z) &= \sum_{i=1}^{n-1} f(z_i), \\
\text{where} \quad
f(z_i) &=
\begin{cases}
1 & \text{if } \operatorname{sign}(z_i) \neq \operatorname{sign}(z_{i+1}), \\
0 & \text{otherwise}.
\end{cases}
\end{aligned}
\end{equation}

\section{On the importance of choosing the right input representations and pooling operator}\label{sec:importance}

\subsection{Influence of input representations and pooling operators}
To assess the influence of input representations and pooling operators on the performance of transformations based on random convolution kernels, we conducted a comparison using 112 selected datasets of the UCR archive~\cite{dau_ucr_2018}. This comparison involved the original MiniRocket and modified versions, which utilized ZC, MPV, MIPV, and LSPV as pooling operators instead of PPV. GMP was excluded, as it was outperformed in virtually all cases. Additionally, we incorporated the first-order difference (DIFF) as an extra input representation.  Classifiers are denoted using a notation $\textbf{PO\_IR}$ depending on the pooling operator $\textbf{PO}$ used ($PO \in \{PPV, ZC, MPV, MIPV, LSPV\}$) and the input representation $\textbf{IR}$ used (MIX denoting the concatenation of the first order difference DIFF and the base representation). We then compared $15$ possible transformations, which are formally defined in Section~\ref{sec:selfrocket:fg}. These transformations correspond to combinations of these 5 potential pooling operators with these 3 possible input representations: the base representation, DIFF, and MIX. It is worth mentioning that we also conducted experiments using multiple pooling operators, obtained by concatenating two or more $PO$, rather than selecting just one as described here. However, our preliminary results suggested that using multiple operators did not yield performances as effective as those discussed in this section.

Table~\ref{fig:TAB35ALLPO} illustrates the mean classification accuracy across $30$ resamples of the first $35$ datasets included in UCR archive for the $15$ possible transforms. For each dataset, the number in bold indicates the highest overall accuracy achieved. Note that the PPV column is also the performance of the original MiniRocket classifier.

\begin{sidewaystable}
    \caption{Mean classification accuracy for 30 resamples of the 35 first datasets for each couple of $IR \times PO$.}
    \footnotesize{
    \begin{tabular}{llllllllllllllll}
        \multirow{2}{*}{DATASET} & PPV & ZC & MPV & MIPV & LSPV & PPV & ZC & MPV & MIPV & LSPV & PPV & ZC & MPV & MIPV & LSPV \\
        & & & & & & DIFF & DIFF & DIFF & DIFF & DIFF & MIX & MIX & MIX & MIX & MIX \\ \hline \\
                 ACSF1 & 82.53 & 81.63 & 82.93 & 69.6 & 85.2 & 79.83 & 78.57 & 81.83 & 69.13 & 84.63 & 82.23 & 82.43 & 82.77 & 70.5 & \textbf{\underline{85.8}} \\ 
        Adiac & 80.09 & 79.63 & 80.47 & 78.88 & 79.97 & 82.35 & 81.35 & 79.28 & 75.31 & 79.36 & \textbf{\underline{82.86}} & 81.76 & 81.9 & 79.06 & 81.34 \\ 
        ArrowHead & 88.19 & 86.67 & 87.71 & 85.56 & 88.63 & 87.96 & 88.04 & 86.04 & 81.05 & 88.59 & 88.53 & 88.34 & 88 & 85.05 & \textbf{\underline{89.14}} \\ 
        Beef & 76.78 & 69.33 & 75.56 & 73.56 & 68.89 & 77.89 & 72.11 & 71.33 & 77.67 & 68.11 & \textbf{\underline{79.78}} & 71.11 & 74.89 & 76.89 & 71.22 \\ 
        BeetleFly & 91.33 & 89.5 & 85.33 & 88.17 & 87.17 & 91.17 & \textbf{\underline{91.83}} & 91 & 88.83 & 87.67 & 91.83 & 90.67 & 88.5 & 91.33 & 88 \\ 
        BirdChicken & 91.83 & 92.17 & 89.5 & 83.33 & 91.5 & 90.17 & \textbf{\underline{92.5}} & 91.83 & 87.17 & 89.67 & 90 & 92.17 & 88.17 & 85 & 90 \\ 
        BME & 99.2 & 99.11 & 99.8 & 99.36 & 92.56 & 98.16 & 94.73 & 99.84 & 99.4 & 92.16 & 99.22 & 98.4 & \textbf{\underline{99.87}} & 99.53 & 93.8 \\ 
        Car & 91.89 & 88.72 & 91.06 & 91.22 & 88.83 & 90.61 & 89.28 & 89.28 & 88.39 & 83.89 & \textbf{\underline{92.61}} & 91.11 & 92 & 91.17 & 89.22 \\ 
        CBF & 99.63 & \textbf{\underline{99.89}} & 99.41 & 96.03 & 99.38 & 83.37 & 75.15 & 80.93 & 79.67 & 81.37 & 99.36 & 99.68 & 99.26 & 94.4 & 98.85 \\ 
        Chinatown & 96.88 & 97 & 96.7 & \textbf{\underline{97.27}} & 96.03 & 95.8 & 94.58 & 94.99 & 96.4 & 94.68 & 96.58 & 96.72 & 96.09 & 97.07 & 96.03 \\ 
        ChlorineConcentration & 75.54 & 75.94 & 71.23 & 75.94 & 73.45 & 78.64 & \textbf{\underline{78.94}} & 77.69 & 78.51 & 75.89 & 78.18 & 78.12 & 77.45 & 78.63 & 76.08 \\ 
        CinCECGTorso & 87.58 & 90.93 & 87.07 & 93.47 & 87.24 & 95.65 & \textbf{\underline{98.17}} & 93 & 96.54 & 86.77 & 92.55 & 96.62 & 90.81 & 95.97 & 89.64 \\ 
        Coffee & 99.88 & 97.98 & \textbf{\underline{100}} & 100 & 100 & 98.81 & 98.45 & 99.88 & 99.64 & 98.57 & 99.76 & 98.81 & 100 & 100 & 100 \\ 
        Computers & 80.19 & 80.11 & 81.4 & 72.67 & 79.41 & 84.45 & 83.83 & 81.85 & 74.77 & 80.47 & \textbf{\underline{84.93}} & 84.75 & 84.53 & 75.13 & 83.04 \\ 
        CricketX & \textbf{\underline{82.46}} & 81.56 & 81.27 & 79.44 & 80.83 & 70.82 & 73.15 & 68.65 & 65.56 & 69.16 & 81.63 & 82.12 & 81.78 & 78.31 & 79.68 \\ 
        CricketY & 84.18 & 83.83 & 82.61 & 79.47 & 81.44 & 68.75 & 69.56 & 67.34 & 65.86 & 67.71 & 83.62 & \textbf{\underline{84.22}} & 82.85 & 80.08 & 80.47 \\ 
        CricketZ & 84.1 & \textbf{\underline{84.52}} & 83.44 & 81.02 & 82.24 & 72.4 & 74.56 & 70.19 & 65.29 & 71.33 & 83.18 & 84.26 & 83.8 & 79.91 & 81.94 \\ 
        Crop & 76.42 & 75.94 & 76.18 & 75.97 & 76.29 & 70.65 & 71.37 & 70.97 & 70.59 & 70.26 & 76.69 & 76.51 & \textbf{\underline{76.94}} & 76.16 & 76.36 \\ 
        DiatomSizeReduction & 94.24 & 91.46 & 94.73 & 95.85 & 95.36 & 93.71 & 91.44 & 94.02 & 95.44 & 92.06 & 94.24 & 91.8 & 94.6 & \textbf{\underline{95.95}} & 95.16 \\ 
        DistalPhalanxOutlineAgeGroup & 79.83 & 80.1 & 79.9 & 80.02 & 79.18 & 80.24 & 80.17 & \textbf{\underline{80.82}} & 80.34 & 79.86 & 79.95 & 80.5 & 79.71 & 80.43 & 80.26 \\ 
        DistalPhalanxOutlineCorrect & 82.6 & 82.86 & 82.84 & 83.02 & 83.29 & \textbf{\underline{84.02}} & 83.41 & 83.31 & 83.79 & 83.13 & 84 & 83.96 & 83.64 & 83.76 & 83.49 \\ 
        DistalPhalanxTW & 69.16 & 69.62 & 69.81 & 69.42 & 69.28 & 69.86 & 69.21 & \textbf{\underline{70.07}} & 68.37 & 68.9 & 69.64 & 69.59 & 69.74 & 68.99 & 69.35 \\ 
        Earthquakes & 74.03 & 73.93 & 74.1 & 75.18 & 73.91 & 74.48 & 74.53 & 74.34 & 74.92 & 74.68 & 74.2 & 74.1 & 74.27 & \textbf{\underline{75.18}} & 74.39 \\ 
        ECG200 & \textbf{\underline{90.03}} & 89.8 & 89.77 & 88.13 & 89.2 & 84.67 & 85.77 & 85.1 & 84.3 & 82.13 & 89.57 & 89.67 & 89.27 & 87.8 & 87.73 \\ 
        ECG5000 & 94.65 & 94.5 & 94.71 & 94.56 & 94.54 & 94.19 & 94.2 & 94.43 & 94.25 & 94.16 & 94.65 & 94.57 & \textbf{\underline{94.74}} & 94.57 & 94.57 \\ 
        ECGFiveDays & 99.05 & \textbf{\underline{99.73}} & 99.66 & 97.46 & 99.54 & 98.78 & 98.19 & 98.96 & 96.9 & 98.36 & 99.18 & 99.56 & 99.71 & 97.24 & 99.66 \\ 
        ElectricDevices & 87.4 & 87.26 & 87.74 & 85.12 & 87.11 & 85.94 & 86 & 85.61 & 79.73 & 84.84 & \textbf{\underline{89.21}} & 88.98 & 88.97 & 86 & 88.15 \\ 
        EOGHorizontalSignal & 83.51 & 80.36 & 84.73 & 80.76 & 83.93 & 80.19 & 79.1 & 80.89 & 73.24 & 76.61 & 85.1 & 82.73 & \textbf{\underline{86.66}} & 82.15 & 85.08 \\ 
        EOGVerticalSignal & 79.71 & 76.07 & 80.52 & 76.76 & 79.25 & 73.51 & 73.3 & 75.78 & 68.48 & 73.12 & 79.53 & 78.61 & \textbf{\underline{81.61}} & 76.57 & 79.31 \\ 
        EthanolLevel & 66.49 & 65.22 & 62.84 & \textbf{\underline{71.87}} & 70.73 & 60.14 & 60.69 & 51.61 & 65.47 & 52.2 & 66.87 & 67 & 62.54 & 71.73 & 69.87 \\ 
        FaceAll & 98.33 & 98.15 & \textbf{\underline{98.62}} & 98.32 & 97.99 & 96.95 & 96.42 & 97.06 & 97.17 & 95.92 & 98.37 & 98.09 & 98.62 & 98.28 & 97.79 \\ 
        FaceFour & \textbf{\underline{93.98}} & 89.36 & 92.42 & 90.11 & 92.8 & 77.99 & 78.6 & 75.42 & 81.44 & 78.67 & 92.61 & 86.06 & 89.73 & 88.56 & 90.45 \\ 
        FacesUCR & 96.9 & 95.95 & 96.7 & \textbf{\underline{97.15}} & 95.84 & 92.45 & 90.47 & 93.1 & 93.67 & 90.98 & 96.56 & 95.31 & 96.61 & 96.77 & 95.17 \\
        FiftyWords & 83.02 & 79.93 & 81.7 & 82.35 & 79.88 & 80.12 & 74.18 & 74.4 & 79.37 & 74.84 & \textbf{\underline{83.66}} & 80.39 & 81.52 & 82.51 & 79.59 \\
    \end{tabular}
    }
    \label{fig:TAB35ALLPO}
\end{sidewaystable}

\begin{figure}[ht]
        \centering
        \includegraphics[width=0.7\textwidth]{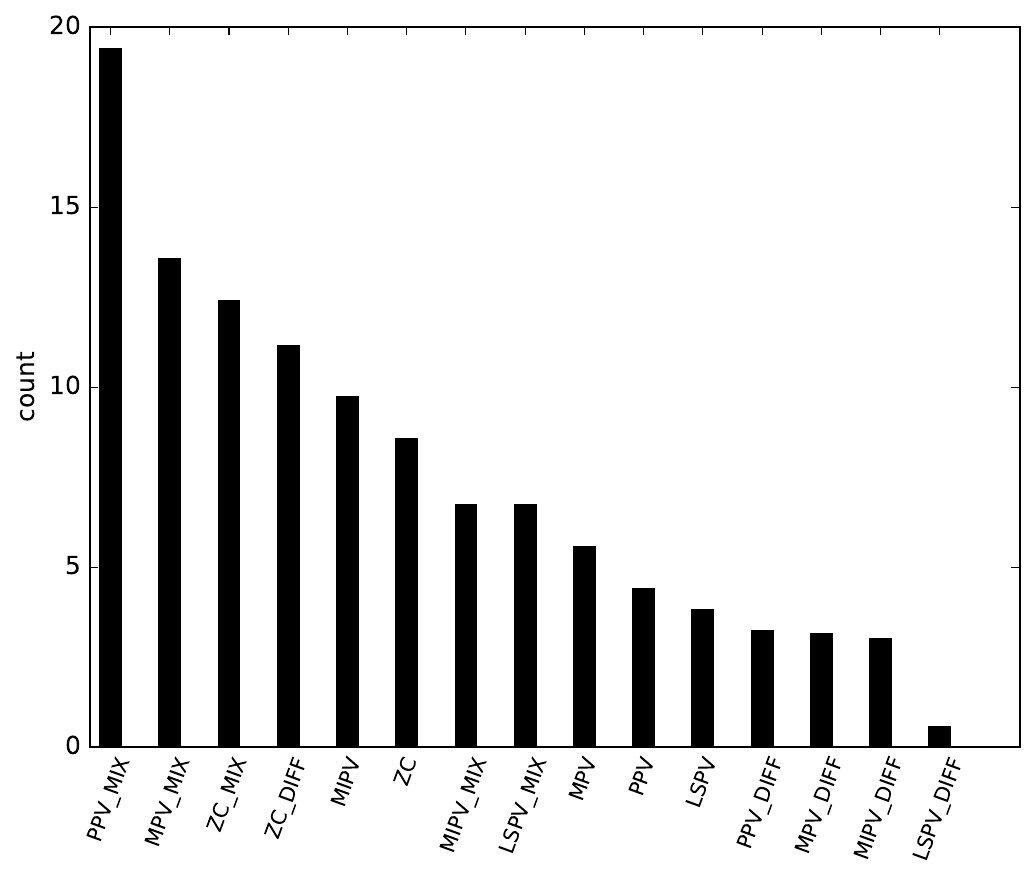}
        \caption{Number of datasets in the UCR archive for which the modified MiniRocket version performs the best}
        \label{img:count_PO}
    \end{figure}

Figure~\ref{img:count_PO} shows the number of datasets in the UCR archive for which the modified MiniRocket version performs the best. If PPV\_MIX is the best transform on average, it is outperformed by others in most cases (93 cases over 112 $\approx 83\%$).

\begin{figure}[t]
        \centering
        \includegraphics[width=0.9\textwidth]{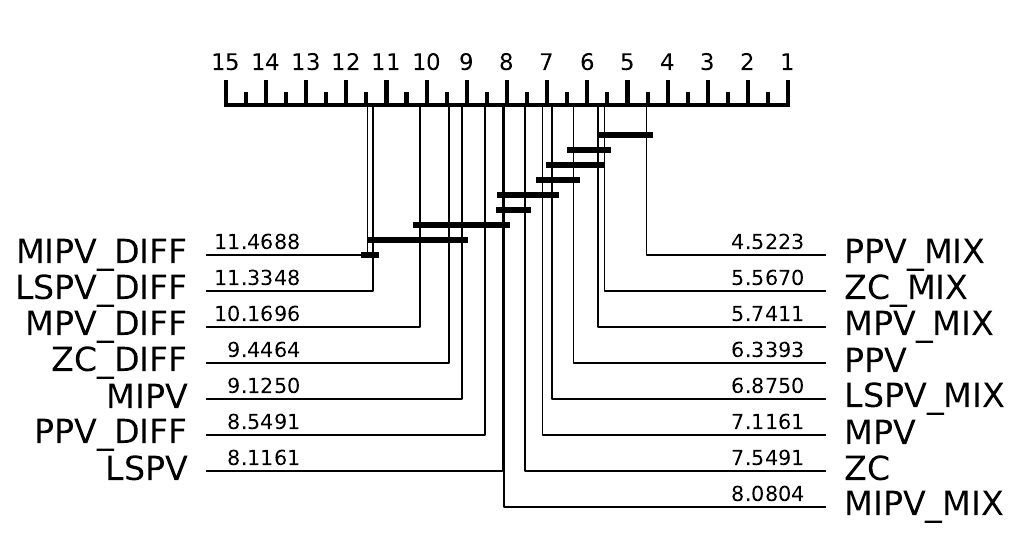}
        \caption{Critical difference diagram of the tested transforms.}
        \label{img:CDD_PO}
    \end{figure}

Figure~\ref{img:CDD_PO} displays the critical difference diagram~\cite{demsar2006statistical} of the 15 possible transforms for the different datasets. In Appendix~\ref{sec:A1}, the distribution of classification accuracy is also provided as box plots in Figure~\ref{fig:Distrib_PO_Detailed}.

Based on these experiments, the combination of input representations and pooling operators that yields the best performance varies across datasets. Therefore, it is beneficial to be able to select the appropriate combination based on the training data. This is the goal of SelF-Rocket, which we present in Section~\ref{sec:selfrocket}. Before that, we examine whether certain dataset properties influence the choice of the most appropriate IR-PO.

\subsection{IR-PO selection and dataset properties}

The only consistent trend we observed concerns time series length. PPV tends to perform best on short time series, whereas ZC consistently outperforms PPV for longer series as shown in Table~\ref{tab:PERFLENTS}. This behavior can be interpreted as follows: since series length increases, ZC benefits from accumulating sign changes that reflect richer temporal dynamics and oscillatory behavior while PPV captures how well the kernel aligns with a global pattern in the time series, making it particularly effective for shorter sequences. This result may also explain MiniRocket's strong performance on the UCR archive, which contains a large proportion of short time series, where PPV-based pooling is particularly effective.\newline

\begin{table}[!ht]
    \centering
      \caption{Impact of time series length on the performance of PO\_MIX over the 112 UCR datasets}
    \begin{tabular}{|l|l|l|l|l|l|l|}
    \hline
        Length  & PPV\_MIX & LSPV\_MIX & MIPV\_MIX & MPV\_MIX & GMP\_MIX & ZC\_MIX \\ \hline
        $< 128$ & \textbf{85.82} & 84.96 & 84.7 & 85.61 & 82.2 & 85.46 \\ \hline
        128-255 & \textbf{96.21} & 94.98 & 95.07 & 96.13 & 94.6 & 95.57 \\ \hline
        256-511 & \textbf{89.75} & 88.12 & 84.2 & 88.95 & 83.39 & 88.37 \\ \hline
        512-767 & 85.75 & 85.03 & 83.1 & 85.27 & 79.99 & \textbf{86.01} \\ \hline
        768-999 & 83.44 & 83.91 & 77.35 & 84.35 & 79.91 & \textbf{84.66} \\ \hline
        $> 999$ & 81.43 & 82.62 & 72.26 & 80.43 & 70.13 & \textbf{82.68} \\ \hline
    \end{tabular}
    \label{tab:PERFLENTS}
\end{table}

\begin{figure*}[ht]
        \centering
        \begin{subfigure}[b]{0.475\textwidth}
            \centering
            \includegraphics[width=\textwidth]{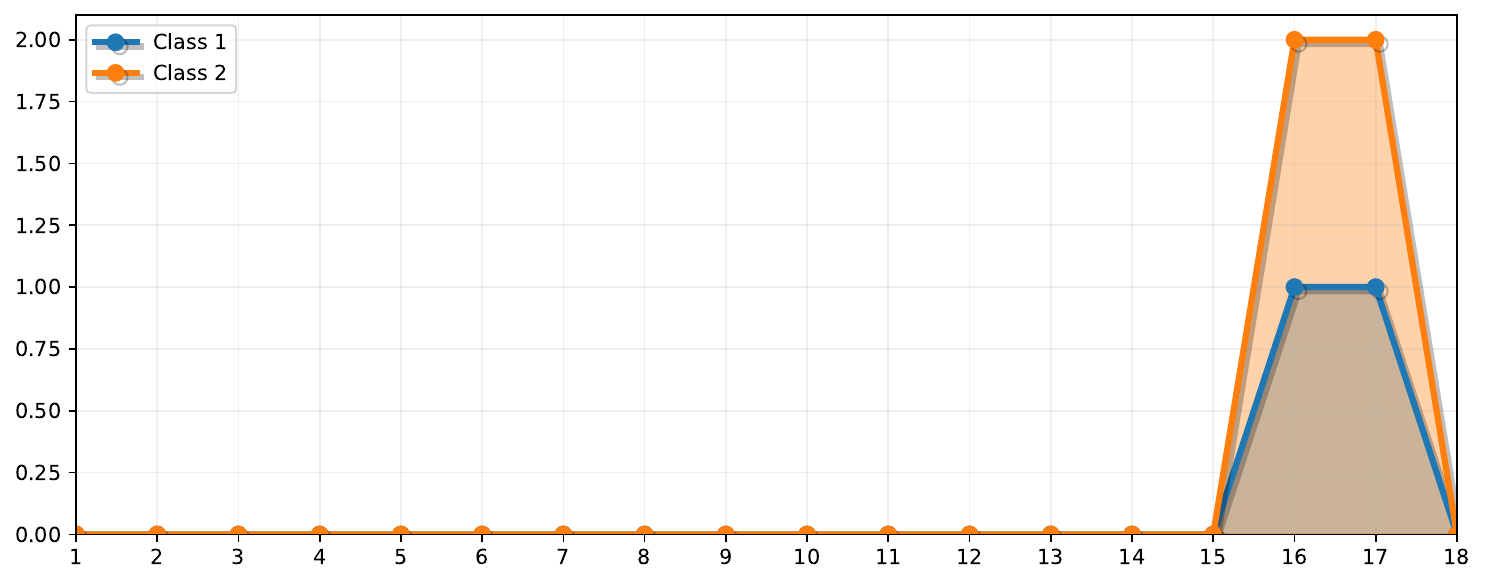}
            \caption[]%
            {{\footnotesize  Scenario \#1: Difference of pattern Vertical Scale}}   
            \label{fig:}
        \end{subfigure}
        \hfill
        \begin{subfigure}[b]{0.475\textwidth}  
            \centering 
            \includegraphics[width=\textwidth]{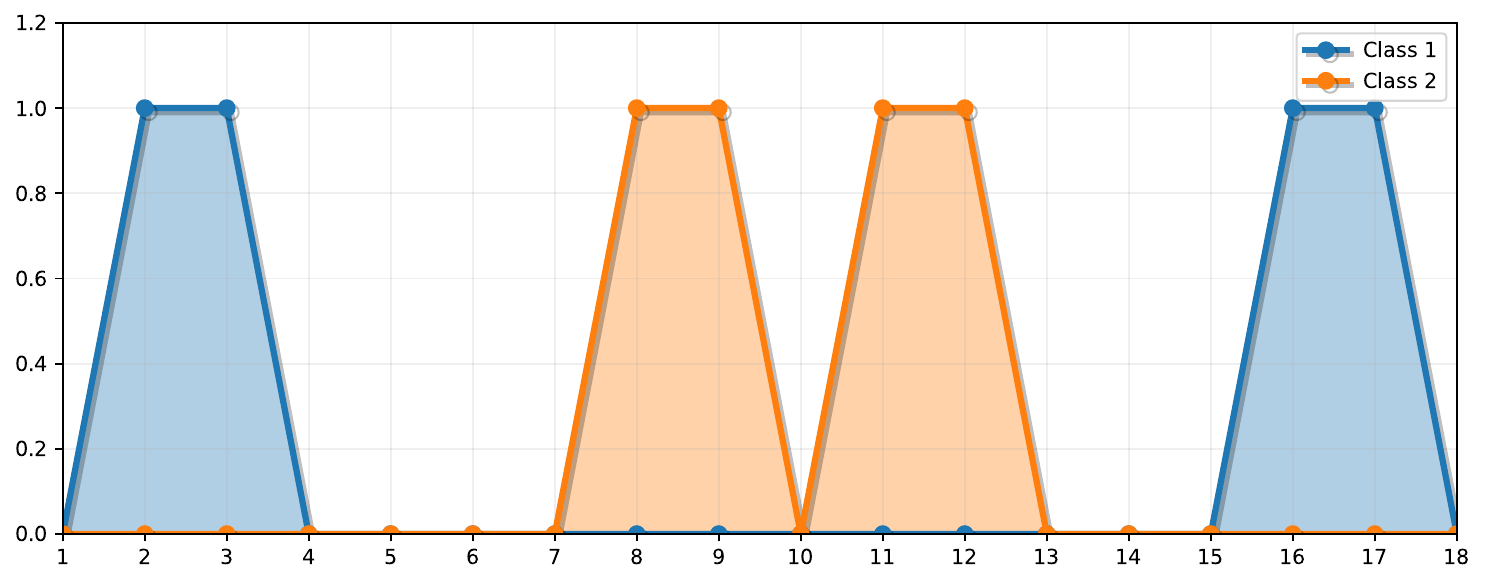}
            \caption[.]%
            {{\footnotesize Scenario \#2: Difference of pattern Mean Position} }  
            \label{fig:}
        \end{subfigure}
        \vskip\baselineskip
        \begin{subfigure}[b]{0.475\textwidth}   
            \centering 
            \includegraphics[width=\textwidth]{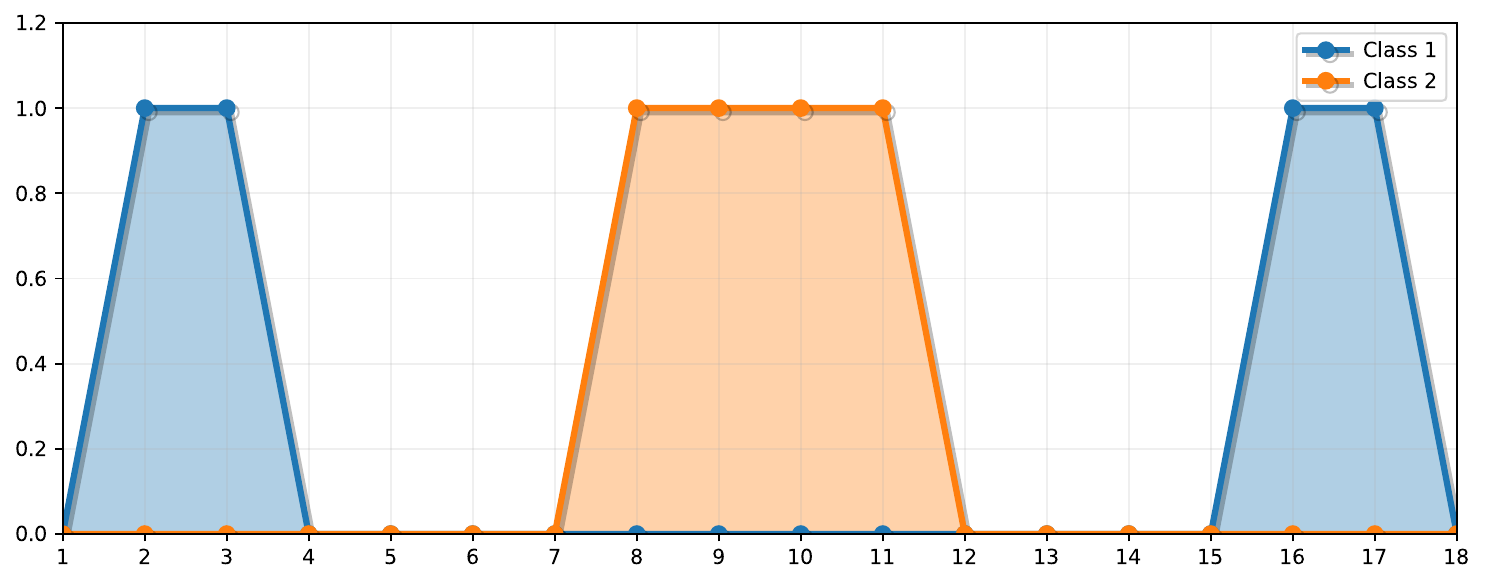}
            \caption[]%
            {{\footnotesize  Scenario \#3: Difference of pattern Horizontal Scale}}    
            \label{fig:}
        \end{subfigure}
        \hfill
        \begin{subfigure}[b]{0.475\textwidth}   
            \centering 
            \includegraphics[width=\textwidth]{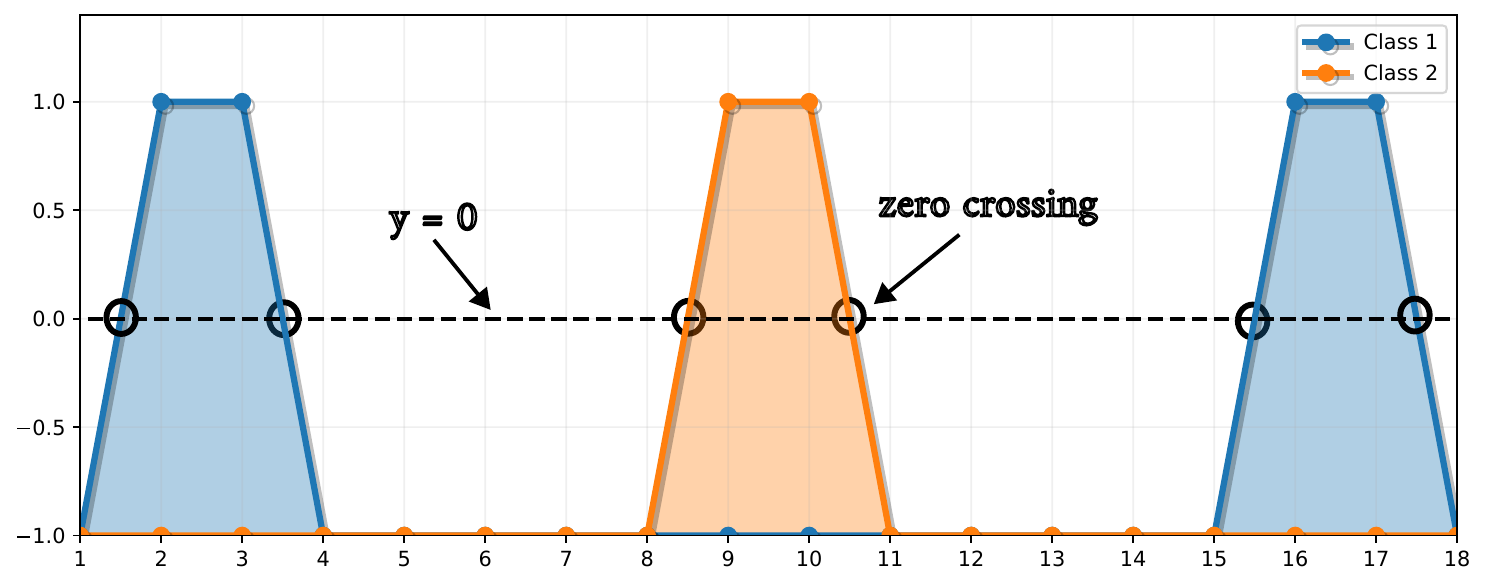}
            \caption[.]%
            {{\footnotesize  Scenario \#4: Difference of pattern Occurences \phantom{this text is invisible}} }  
            \label{fig:}
        \end{subfigure}
        \caption[.]
        {\small Four synthetic activation map scenarios illustrating class-dependent variations of a discriminative pattern after convolution}
        \label{fig:SYNTHCV}
    \end{figure*}

We also investigated the ability of the pooling operators MPV, MIPV, LSPV, PPV and ZC to distinguish between two classes based on the activation maps obtained after convolution with a kernel.
Figure~\ref{fig:SYNTHCV} illustrates four synthetic activation map scenarios with class-dependent variations: differences in vertical scale, mean position, horizontal scale, and pattern occurrences.
Table~\ref{tab:SYNTHBESTPO} shows the most effective pooling operators for distinguishing the two classes in each of the four previously defined scenarios.    

\begin{table}[ht]
  \centering
  \caption{Comparison of class-wise pooling operator values and their differences for Scenarios \#1–4}
\begin{tabular}{|c|*{4}{c} c|}
  \hline
   & PPV & MPV & MIPV & LSPV & ZC  \\
  \hline
   \multicolumn{6}{|l|}{Scenario \#1}\\
  \hline
  Class 1  & 0.11  & 1   &  16.5  & 2 & 0  \\
  Class 2 & 0.11   & 2   & 16.5   & 2   & 0 \\
  $\Delta$ & 0   & \textbf{1} & 0 & 0  & 0\\
  \hline
  \multicolumn{6}{|l|}{Scenario \#2}\\
  \hline
  Class 1  & 0.22  & 1   &   9.5  & 2 & 0  \\
  Class 2 & 0.22   & 1   & 10   & 2   & 0 \\
  $\Delta$ & 0  & 0 & \textbf{0.5} & 0  & 0\\
\hline
  \multicolumn{6}{|l|}{Scenario \#3}\\
  \hline
  Class 1  & 0.22  & 1   &   9.5  & 2 & 0  \\
  Class 2 & 0.22   & 1   & 9.5   & 4   & 0 \\
  $\Delta$ &  0   & 0 & 0 & \textbf{2}  & 0\\
\hline
 \multicolumn{6}{|l|}{Scenario \#4}\\
  \hline
  Class 1  & 0.22  & 1   &   9.5  & 2 & 4  \\
  Class 2 & 0.11   & 1   & 9.5  & 2   & 2 \\
  $\Delta$ &  \textbf{0.11}  & 0 & 0 & 0  & \textbf{2}\\
  \bottomrule
\end{tabular}
\label{tab:SYNTHBESTPO}
\end{table}

Overall, MPV performs best when the classes differ in amplitude (Scenario~\#1). MIPV, in contrast, achieves better performance when the average location of discriminative patterns differs from one class to another (scenario \#2). LSPV performs better when the discriminative pattern exhibits class-dependent variations in duration (scenario \#3). Finally, PPV and ZC act as versatile operators, performing well when the pattern appears at different frequencies depending on the class (scenario \#4).

These observations also indicate that the pooling operator that yields the best performance varies across datasets and class structures. 

\section{SelF-Rocket}\label{sec:selfrocket}

\begin{figure}[!t]
    \centering
    \resizebox{1\textwidth}{!}{%
    \input{images/Archi.tikz}
    }%
    \caption{SelF-Rocket Training Architecture}
    \label{fig:Architecture}
 \end{figure}
 
Following the experimental results obtained in the previous section, we investigate a new variant of ROCKET, called SelF-Rocket (\textbf{Sel}ected \textbf{F}eatures Rocket), that aims to dynamically select the best couple of input representations and pooling operator during the training process. The latter, illustrated in Figure~\ref{fig:Architecture}, involves three main stages:
\begin{enumerate}
    \item The feature set generation step, detailed in Section~\ref{sec:selfrocket:fg};
    \item The feature selection step, exposed in Section~\ref{sec:selfrocket:fsm};
    \item And the classification step, presented in Section~\ref{sec:selfrocket:c}. 
\end{enumerate}

SelF-Rocket operates similarly to MiniRocket or MultiRocket, the main difference being the inclusion of a feature set selection step. As previously observed in Section~\ref{sec:importance}, the optimal set of features can differ from one dataset to another. Therefore, while using a single feature type like PPV, as in MiniRocket, or a fixed set of feature types such as \{PPV, LSPV, MIPV, MPV\}, as in MultiRocket, can yield generally good results, it may not be optimal for every specific case. To tackle this problem, we have incorporated a wrapper-based Feature Selection Module with the objective to select the most appropriate set of features for each classification problem.

\subsection{Feature Generation}\label{sec:selfrocket:fg}





Let $\textbf{IR}$ be a set of Input Representations, $\textbf{PO}$ a set of Pooling Operators, $\textbf{MK}$ the set of all kernels (containing every kernel/dilation combination) generated by MiniRocket and $b_\kappa$ the bias associated to a kernel $\kappa \in \textbf{MK}$ (Section~\ref{sec:relatedworks}).

In the implementation evaluated in this article we consider
\begin{equation}
    \textbf{IR} = \{I, DIFF\} 
\end{equation}
and
\begin{equation}
    \textbf{PO} = \{PPV, ZC, MPV, MIPV, LSPV\},
\end{equation}
where $I$ is the identity function ($I(X)=X$ for all time series $X$) and $DIFF$ is the first order difference (Equation~\ref{eqn:difference}).\newline 

Let $f$ be a feature set generation function defined as:
\begin{equation}
    f(X, A, p) = \{p(r(X) \circledast \kappa - b_\kappa) \, | \, \kappa \in \textbf{MK}, r \in A\},
\end{equation}
where $X$ is a time series, $A$ a subset of the power set of \textbf{IR} (minus the empty set), i.e. $A\subseteq 2^{\textbf{IR}}\setminus\{\emptyset\}$, and $p$ a pooling operator, i.e. $p \in \textbf{PO}$.

Thus, following this formalization, the original MiniRocket extracts features using the parametrization $f(X, \{I\}, PPV)$, for any time series $X$.

The Cartesian product $2^{\textbf{IR}}\setminus\{\emptyset\} \times \textbf{PO}$ induces a set of such parameterized functions; in this implementation, $|2^{IR}\setminus\{\emptyset\}| = 3$ and $|PO| = 5$, leading to $N = 3 \times 5 = 15$ possible parametrizations.

Let $\textbf{FV}$ be the generated Feature Vector.
In the following, feature generation outputs are denoted as in Section~\ref{sec:importance} for readability reasons:
\begin{itemize}
\item p for $f(X, \{I\}, p)$,
\item p\_DIFF for  $f(X, \{DIFF \}, p)$,
\item and p\_MIX for  $f(X, \{I, DIFF \}, p)$,
\end{itemize}
for all $p \in \textbf{PO}$.

\subsection{Feature Selection}\label{sec:selfrocket:fsm}

Summarized in Figure~\ref{img:SROV} and Algorithm~\ref{alg:FSM}, the Feature Selection Module relies on stratified train/test split methods to generate new train and validation sets derived from the original train set. Depending on the size of that dataset, we use either a stratified $k$-fold repeated $nr$ times or a stratified shuffle split with $k \times nr$ splits. A stratified shuffle split is used if the dataset contains at least a sufficiently large number of data (denoted by $\bm{mds}$ in Algorithm~\ref{alg:FSM}), it should be noted that this value has a direct impact on the time available for finding the best IR-PO combination by limiting the maximum size of each train and validation set, and thus the time available for this task. This strategy ensures that, as the size of the dataset increases, the learning and validation sets will vary more significantly from one voter to another. It therefore allows more reliable assessments of the performance of the different combinations. These sets are then used to train $k \times nr$ mini-classifiers for each IR-PO combination. 

Subsequently, the selection of the optimal IR-PO combination across all the original train set splits is made using a highest median voting system, which is validated by the Algorithm~\ref{alg:VOTVAL} prior to its use as the final set of features for the linear classifier. Using the highest median to select an IR-PO as an alternative to the highest mean is preferable as it is more robust to extreme values (e.g. lucky runs) that may occur. The idea of the Algorithm~\ref{alg:VOTVAL} is to check if the selected IR-PO combination is sufficiently supported by the voters, otherwise a default combination is chosen. Its aim is to avoid choosing an IR-PO combination that has poor generalisability, especially with small datasets. In our implementation, Algorithm~\ref{alg:VOTVAL} checks if the selected IR-PO combination is part of the top values for each voter (e.g. in the top $4$ for each voter) with a certain degree of flexibility (e.g. at $\geq0.95$, implemented as threshold $\bm{vote_{thresh}}$ in Algorithm~\ref{alg:VOTVAL}), otherwise the selected IR-PO combination is replaced by a default one (PPV\_MIX or ZC\_MIX, depending on the length of the time series).

\begin{algorithm}[hbt!]
\small
\caption{Feature Selection Module}\label{alg:FSM}
\SetKwInOut{Input}{Input}
\SetKwInOut{Output}{Output}
\SetKwInOut{Parameters}{Parameters}

\Input{Transformed train Feature Vector $\bm{FV}$ \newline Training set class $\bm{y_{train}}$ }
\Parameters{Number of folds $\bm{k}$  \newline Number of features per mini-classifier $\bm{f}$ \newline Number of runs $\bm{nr}$ \newline Maximum dataset size $\bm{mds}$}
\Output{Optimal set of features $\textbf{S}$}

      \uIf{$length(y_{train}) \leq  mds$}{
    $\text{splt} \gets \text{RepeatedStratifiedKFold}(n\_splits = k, \text{ } n\_repeat= nr)$ \;
  }
  \Else{
    $\text{splt} \gets \text{StratifiedShuffleSplit}(n\_splits = k \times nr, \text{ } train\_size = int(mds/2), \text{ } test\_size = int(mds/2))$ \;
  }

$\text{performances} \gets \text{List()}$ ;

\For{$l\in [0,1,\ldots,k \times nr - 1]$ }{

    $ind_{train} \gets \text{splt[l][0]}$ \tcp*{New Train set indices}
    
    $ind_{val} \gets \text{splt[l][1]}$ \tcp*{Validation set indices}

    \For{$t\in [0,1,\ldots,|FV| - 1]$ }{

        $\text{classifier} \gets \text{RidgeClassifier}()$;

        \tcp{Selection of $f$ random features indices}
        $ ind_{feats} \gets \text{Random}([0,1,\ldots,|FV[t]| - 1],f$)  ;
        
        $ feats_{train} \gets FV[t][ind_{train}][:,ind_{feats}] $ ;
        
        $ feats_{val} \gets FV[t][ind_{val}][:,ind_{feats}] $ ;
        
        $ y_{train} \gets y_{train}[ind_{train}]$ ;

        $ y_{val} \gets y_{train}[ind_{val}]$ ;

        $\text{classifier.train(}feats_{train},y_{train}\text{)}$ ;

        $ y_{pred} \gets \text{classifier.predict(}feats_{val}\text{)}$ ;

        $\text{performances.add(AccuracyScore(}y_{val}, y_{pred}\text{))}$ ;
        
    }
}
$idx_{hvms} \gets \text{HighestMedianVoting(performances)}$ ;

$idx_{final} \gets \text{VoteValidation(performances, }idx_{hvms},\text{ } nb_{voters}=k\times nr\text{)}$ ;

$\textbf{S} \gets FV[idx_{final}]$ ;

\Return \textbf{S}

\end{algorithm}

\begin{algorithm}[hbt!]
\small
\caption{Vote Validation}\label{alg:VOTVAL}
\SetKwInOut{Input}{Input}
\SetKwInOut{Output}{Output}
\SetKwInOut{Parameters}{Parameters}

\Input{Performance Vector $\bm{PV}$ \newline Index of the chosen IR-PO combination $\bm{idx_{vote}}$ \newline Number of voters $\bm{nb_{vot}}$ \newline Time series length $\bm{len_{ts}}$}
\Parameters{Index of PPV\_MIX $\bm{idx_{ppvmix}}$ \newline Index of ZC\_MIX $\bm{idx_{zcmix}}$ \newline Value of the Vote threshold $\bm{vote_{thresh}}$ \newline Value of the length Threshold $\bm{len_{thresh}}$\newline Top considered $\bm{top}$}
\Output{Final index selected $\bm{idx_{final}}$}

$\text{counter} \gets \text{List()}$ ;

\tcp{Check if the selected IR-PO is part of the top values for each voter}

\For{$vot\in [0,1,\ldots,nb_{vot} - 1]$ }{
  $ \text{counter.Add(IsInTopValue}(PV[vot][idx_{vote}],PV[vot], top\text{))} $ ;
}
\uIf{$mean(\text{counter})\geq vote_{thresh}$}{
    $idx_{final} \gets idx_{vote}$ \;
  }
  \Else{
    \uIf{$len_{ts}\geq {len_{thresh}}$}{
    $idx_{final} \gets idx_{zcmix}$ \;
  }
  \Else{
    $idx_{final} \gets idx_{ppvmix}$ \;
  }
    
} 
  
\Return $idx_{final}$
\end{algorithm}

\begin{figure}[!t]
    \centering
    \resizebox{1\textwidth}{!}{%
    \input{images/SROV.tikz}
    }%
    \caption{SelF-Rocket Feature Selection Module Overview}
    \label{img:SROV}
\end{figure}

\subsection{Classification}\label{sec:selfrocket:c}

SelF-Rocket employs the same algorithm for the mini-classifiers embedded within the Feature Selection Module and the end-stage classifier, namely the Ridge classifier, as proposed by Dempster et al.~\cite{dempster2020rocket}. This classifier is preferable to stochastic descent methods, such as logistic regression, when the number of features exceeds the number of training examples and when working with small datasets, e.g. when there are fewer than 10,000 examples.

\section{Experiments}\label{sec:experiments}
In this section, the performances of SelF-Rocket and Hydra + SelF-Rocket (concatenation of the features of Hydra and SelF-Rocket) are evaluated. We show that SelF-Rocket is as accurate as MultiRocket, despite the use of only one pooling operator at a time. We also investigate the influence of the main parameters of SelF-Rocket on its performance.

\subsection{Experimental settings}

To guarantee the reproducibility of our experiments and a fair comparison with other TSC algorithms, the proposed method is evaluated on the same same identical 112 datasets of the UCR Time Series Archive \cite{dau_ucr_2018} and the precise same 30 train/test resamples for each of those datasets~\cite{dempster2020rocket,dempster2023hydra,dempster2024quant,tan2022multirocket,tan2025proximity,middlehurst_bake_2024}. The 112 datasets are those from the 128 original ones that did not contain missing values or variable time series length. 

The evaluation of a TSC method on UCR may be limited~\cite{dau_ucr_2018,hu2016classification} (for example, the data are pre-processed and of the same length), nevertheless, as described above, it currently remains the standard for benchmarking TSC methods.

Note that we did not compute the results for other methods shown in Figures \ref{fig:ARP}, \ref{fig:MCM} and \ref{fig:SRALL}, but used the results\footnote{\url{https://github.com/time-series-machine-learning/tsml-eval/tree/main/results/classification/Univariate}} available in~\cite{middlehurst_bake_2024}.

The original implementations of MiniRocket\footnote{\url{https://github.com/angus924/minirocket}} and MultiRocket\footnote{\url{https://github.com/ChangWeiTan/MultiRocket}} were used as baseline. Our algorithm is implemented in standard Python 3.11.4. \newline

In order to compare the classification performance with other TSC algorithms, we use the critical difference diagrams that display mean ranks of each method across all datasets with the horizontal cliques indicating that there is no statistically significant difference between those methods, the Multiple Comparison Matrix (MCM) \cite{ismailfawaz2023approachmultiplecomparisonbenchmark} with heatmap color representing mean differences in score, and the pairwise scatter plots of test accuracy summarizing the number of win/draw/loss between 2 classifiers across all datasets. \newline

All experiments were conducted on a cluster using Ubuntu 24.04.1 LTS with an Intel(R) Xeon(R) Gold 6434 and 250GB of RAM. \newline

A sensitivity analysis of the main parameters of Self-Rocket is given in Section~\ref{sec:sensitivity}. We have chosen as default parameters for SelF-Rocket, a number of folds of $k=2$, a number of features per mini-classifier of $f=2500$, a number of runs of $nr=10$, and a validation vote with a top $5$ and a threshold of $0.9$. The others parameters of SelF-Rocket, i.e. number of kernels, padding, dilation, bias, length of kernel, values inside the kernel, remain the same as the default parameters of MiniRocket.

\subsection{Comparison with other TSC methods}

We compare the performances of SelF-Rocket and Hydra + SelF-Rocket with 11 TSC methods: 1NN-DTW, Elastic Ensemble (EE)~\cite{lines2015time}, Shapelet transform classifier (STC)~\cite{lines2012shapelet}, HIVE-COTE v2.0~\cite{middlehurst2021hive}, H-Inceptiontime~\cite{ismail2022deep}, MultiRocket~\cite{tan2022multirocket}, Hydra, Hydra+MultiRocket~\cite{dempster2023hydra}, MiniRocket~\cite{dempster2021minirocket},
ROCKET~\cite{dempster2020rocket} and QUANT~\cite{dempster2024quant}. \newline
\begin{figure}
    \centering
    \includegraphics[width=0.8\linewidth]{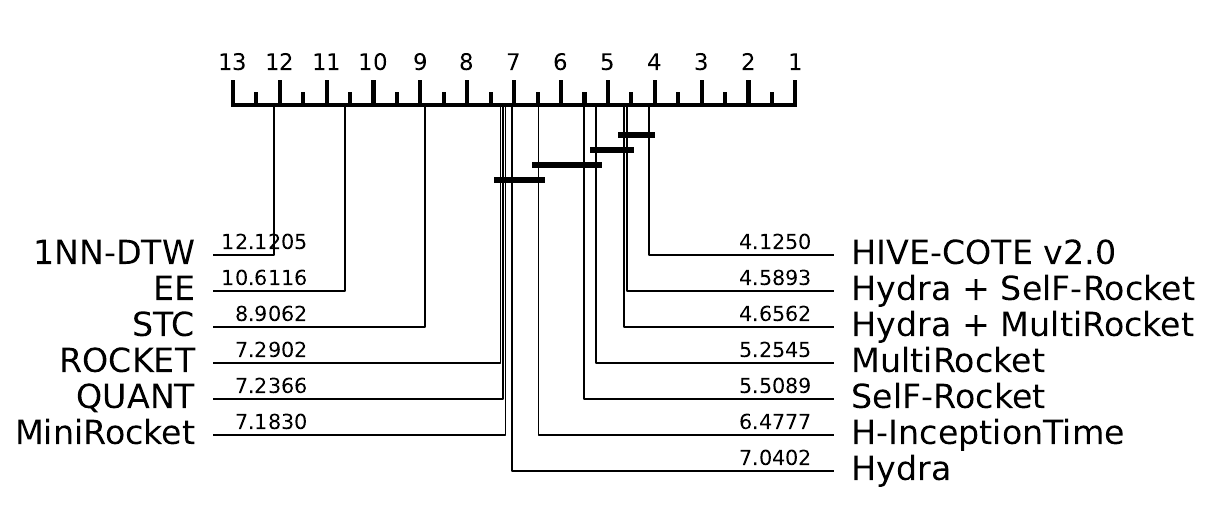}
    \caption{Averaged ranked performance for SelF-Rocket and 11 other classifiers.}
    \label{fig:ARP}
\end{figure}

\begin{figure}
\tiny
\sffamily
\begin{center}
\begin{tabular}{cccccccccc}
Mean-Accuracy & \shortstack{HC2 \\ 0.8914} & \shortstack{H + SelF-R \\ 0.8865} & \shortstack{SelF-R \\ 0.8848} & \shortstack{H + MR \\ 0.8840} & \shortstack{MultiR \\ 0.8814} & \shortstack{H-IT \\ 0.8761} & \shortstack{MiniR \\ 0.8744} & \shortstack{QUANT \\ 0.8671} \\[1ex]
\shortstack{HC2 \\ 0.8914} & \cellcolor[rgb]{0.8674,0.8644,0.8626}\shortstack{\rule{0em}{3ex} Mean-Difference \\ r$>$c / r=c / r$<$c \\ Wilcoxon p-value} & \cellcolor[rgb]{0.9383,0.8089,0.7412}\shortstack{\rule{0em}{3ex} 0.0049 \\ 63 / 3 / 46 \\ 0.2140} & \bfseries \cellcolor[rgb]{0.9546,0.7791,0.6925}\shortstack{\rule{0em}{3ex} 0.0066 \\ 70 / 3 / 39 \\ 0.0010} & \cellcolor[rgb]{0.9595,0.767,0.6741}\shortstack{\rule{0em}{3ex} 0.0074 \\ 59 / 7 / 46 \\ 0.1046} & \bfseries \cellcolor[rgb]{0.9685,0.7158,0.6061}\shortstack{\rule{0em}{3ex} 0.0100 \\ 67 / 6 / 39 \\ 0.0051} & \bfseries \cellcolor[rgb]{0.9513,0.5788,0.4594}\shortstack{\rule{0em}{3ex} 0.0153 \\ 73 / 4 / 35 \\ 0.0007} & \bfseries \cellcolor[rgb]{0.9368,0.5327,0.4181}\shortstack{\rule{0em}{3ex} 0.0170 \\ 84 / 5 / 23 \\  $\leq$ 1e-04} & \bfseries \cellcolor[rgb]{0.8204,0.2868,0.2452}\shortstack{\rule{0em}{3ex} 0.0243 \\ 81 / 5 / 26 \\  $\leq$ 1e-04} \\[1ex]
\shortstack{H + SelF-R \\ 0.8865} & \cellcolor[rgb]{0.7727,0.839,0.9493}\shortstack{\rule{0em}{3ex} -0.0049 \\ 46 / 3 / 63 \\ 0.2140} & \cellcolor[rgb]{0.8674,0.8644,0.8626}\shortstack{\rule{0em}{3ex} -} & \bfseries \cellcolor[rgb]{0.8959,0.8499,0.8235}\shortstack{\rule{0em}{3ex} 0.0017 \\ 70 / 6 / 36 \\ 0.0016} & \cellcolor[rgb]{0.9095,0.8394,0.8003}\shortstack{\rule{0em}{3ex} 0.0025 \\ 51 / 4 / 57 \\ 0.8560} & \cellcolor[rgb]{0.9409,0.8056,0.7352}\shortstack{\rule{0em}{3ex} 0.0051 \\ 59 / 5 / 48 \\ 0.1100} & \bfseries \cellcolor[rgb]{0.9692,0.7058,0.5937}\shortstack{\rule{0em}{3ex} 0.0104 \\ 68 / 3 / 41 \\ 0.0112} & \bfseries \cellcolor[rgb]{0.9677,0.663,0.5443}\shortstack{\rule{0em}{3ex} 0.0121 \\ 84 / 4 / 24 \\  $\leq$ 1e-04} & \bfseries \cellcolor[rgb]{0.9058,0.4552,0.3553}\shortstack{\rule{0em}{3ex} 0.0194 \\ 76 / 4 / 32 \\  $\leq$ 1e-04} \\[1ex]
\shortstack{SelF-R \\ 0.8848} & \bfseries \cellcolor[rgb]{0.7339,0.82,0.9707}\shortstack{\rule{0em}{3ex} -0.0066 \\ 39 / 3 / 70 \\ 0.0010} & \bfseries \cellcolor[rgb]{0.8353,0.8605,0.899}\shortstack{\rule{0em}{3ex} -0.0017 \\ 36 / 6 / 70 \\ 0.0016} & \cellcolor[rgb]{0.8674,0.8644,0.8626}\shortstack{\rule{0em}{3ex} -} & \bfseries \cellcolor[rgb]{0.8796,0.8582,0.8458}\shortstack{\rule{0em}{3ex} 0.0008 \\ 42 / 4 / 66 \\ 0.0112} & \cellcolor[rgb]{0.9194,0.8313,0.7829}\shortstack{\rule{0em}{3ex} 0.0033 \\ 45 / 4 / 63 \\ 0.3279} & \cellcolor[rgb]{0.9659,0.7401,0.6371}\shortstack{\rule{0em}{3ex} 0.0087 \\ 65 / 4 / 43 \\ 0.0632} & \bfseries \cellcolor[rgb]{0.9692,0.7058,0.5937}\shortstack{\rule{0em}{3ex} 0.0104 \\ 74 / 7 / 31 \\  $\leq$ 1e-04} & \bfseries \cellcolor[rgb]{0.9294,0.5123,0.4007}\shortstack{\rule{0em}{3ex} 0.0176 \\ 77 / 4 / 31 \\  $\leq$ 1e-04} \\[1ex]
\shortstack{H + MR \\ 0.8840} & \cellcolor[rgb]{0.719,0.812,0.9777}\shortstack{\rule{0em}{3ex} -0.0074 \\ 46 / 7 / 59 \\ 0.1046} & \cellcolor[rgb]{0.8181,0.8556,0.9146}\shortstack{\rule{0em}{3ex} -0.0025 \\ 57 / 4 / 51 \\ 0.8560} & \bfseries \cellcolor[rgb]{0.8514,0.8631,0.8811}\shortstack{\rule{0em}{3ex} -0.0008 \\ 66 / 4 / 42 \\ 0.0112} & \cellcolor[rgb]{0.8674,0.8644,0.8626}\shortstack{\rule{0em}{3ex} -} & \bfseries \cellcolor[rgb]{0.9095,0.8394,0.8003}\shortstack{\rule{0em}{3ex} 0.0026 \\ 68 / 9 / 35 \\ 0.0003} & \cellcolor[rgb]{0.9616,0.758,0.6618}\shortstack{\rule{0em}{3ex} 0.0079 \\ 66 / 4 / 42 \\ 0.0644} & \bfseries \cellcolor[rgb]{0.9682,0.7208,0.6123}\shortstack{\rule{0em}{3ex} 0.0096 \\ 80 / 5 / 27 \\  $\leq$ 1e-04} & \bfseries \cellcolor[rgb]{0.9368,0.5327,0.4181}\shortstack{\rule{0em}{3ex} 0.0169 \\ 75 / 5 / 32 \\  $\leq$ 1e-04} \\[1ex]
\shortstack{MultiR \\ 0.8814} & \bfseries \cellcolor[rgb]{0.662,0.7755,0.9939}\shortstack{\rule{0em}{3ex} -0.0100 \\ 39 / 6 / 67 \\ 0.0051} & \cellcolor[rgb]{0.768,0.837,0.9525}\shortstack{\rule{0em}{3ex} -0.0051 \\ 48 / 5 / 59 \\ 0.1100} & \cellcolor[rgb]{0.805,0.8517,0.9262}\shortstack{\rule{0em}{3ex} -0.0033 \\ 63 / 4 / 45 \\ 0.3279} & \bfseries \cellcolor[rgb]{0.8181,0.8556,0.9146}\shortstack{\rule{0em}{3ex} -0.0026 \\ 35 / 9 / 68 \\ 0.0003} & \cellcolor[rgb]{0.8674,0.8644,0.8626}\shortstack{\rule{0em}{3ex} -} & \cellcolor[rgb]{0.9434,0.8023,0.7292}\shortstack{\rule{0em}{3ex} 0.0054 \\ 63 / 4 / 45 \\ 0.1944} & \bfseries \cellcolor[rgb]{0.9564,0.7751,0.6864}\shortstack{\rule{0em}{3ex} 0.0070 \\ 80 / 5 / 27 \\  $\leq$ 1e-04} & \bfseries \cellcolor[rgb]{0.9594,0.6103,0.4894}\shortstack{\rule{0em}{3ex} 0.0143 \\ 75 / 5 / 32 \\  $\leq$ 1e-04} \\[1ex]
\shortstack{H-IT \\ 0.8761} & \bfseries \cellcolor[rgb]{0.5326,0.6698,0.9904}\shortstack{\rule{0em}{3ex} -0.0153 \\ 35 / 4 / 73 \\ 0.0007} & \bfseries \cellcolor[rgb]{0.6514,0.7681,0.9959}\shortstack{\rule{0em}{3ex} -0.0104 \\ 41 / 3 / 68 \\ 0.0112} & \cellcolor[rgb]{0.6882,0.7932,0.988}\shortstack{\rule{0em}{3ex} -0.0087 \\ 43 / 4 / 65 \\ 0.0632} & \cellcolor[rgb]{0.7087,0.8057,0.9811}\shortstack{\rule{0em}{3ex} -0.0079 \\ 42 / 4 / 66 \\ 0.0644} & \cellcolor[rgb]{0.7634,0.8351,0.9557}\shortstack{\rule{0em}{3ex} -0.0054 \\ 45 / 4 / 63 \\ 0.1944} & \cellcolor[rgb]{0.8674,0.8644,0.8626}\shortstack{\rule{0em}{3ex} -} & \cellcolor[rgb]{0.8959,0.8499,0.8235}\shortstack{\rule{0em}{3ex} 0.0017 \\ 58 / 4 / 50 \\ 0.5043} & \cellcolor[rgb]{0.967,0.7357,0.6309}\shortstack{\rule{0em}{3ex} 0.0090 \\ 61 / 5 / 46 \\ 0.0502} \\[1ex]
\shortstack{MiniR \\ 0.8744} & \bfseries \cellcolor[rgb]{0.4946,0.633,0.979}\shortstack{\rule{0em}{3ex} -0.0170 \\ 23 / 5 / 84 \\  $\leq$ 1e-04} & \bfseries \cellcolor[rgb]{0.6085,0.7357,0.9994}\shortstack{\rule{0em}{3ex} -0.0121 \\ 24 / 4 / 84 \\  $\leq$ 1e-04} & \bfseries \cellcolor[rgb]{0.6514,0.7681,0.9959}\shortstack{\rule{0em}{3ex} -0.0104 \\ 31 / 7 / 74 \\  $\leq$ 1e-04} & \bfseries \cellcolor[rgb]{0.6673,0.7792,0.993}\shortstack{\rule{0em}{3ex} -0.0096 \\ 27 / 5 / 80 \\  $\leq$ 1e-04} & \bfseries \cellcolor[rgb]{0.729,0.8175,0.9732}\shortstack{\rule{0em}{3ex} -0.0070 \\ 27 / 5 / 80 \\  $\leq$ 1e-04} & \cellcolor[rgb]{0.8353,0.8605,0.899}\shortstack{\rule{0em}{3ex} -0.0017 \\ 50 / 4 / 58 \\ 0.5043} & \cellcolor[rgb]{0.8674,0.8644,0.8626}\shortstack{\rule{0em}{3ex} -} & \cellcolor[rgb]{0.9595,0.767,0.6741}\shortstack{\rule{0em}{3ex} 0.0073 \\ 60 / 5 / 47 \\ 0.0953} \\[1ex]
\shortstack{QUANT \\ 0.8671} & \bfseries \cellcolor[rgb]{0.3286,0.4397,0.8696}\shortstack{\rule{0em}{3ex} -0.0243 \\ 26 / 5 / 81 \\  $\leq$ 1e-04} & \bfseries \cellcolor[rgb]{0.4358,0.5707,0.9517}\shortstack{\rule{0em}{3ex} -0.0194 \\ 32 / 4 / 76 \\  $\leq$ 1e-04} & \bfseries \cellcolor[rgb]{0.4785,0.6166,0.9727}\shortstack{\rule{0em}{3ex} -0.0176 \\ 31 / 4 / 77 \\  $\leq$ 1e-04} & \bfseries \cellcolor[rgb]{0.4946,0.633,0.979}\shortstack{\rule{0em}{3ex} -0.0169 \\ 32 / 5 / 75 \\  $\leq$ 1e-04} & \bfseries \cellcolor[rgb]{0.5597,0.6948,0.9961}\shortstack{\rule{0em}{3ex} -0.0143 \\ 32 / 5 / 75 \\  $\leq$ 1e-04} & \cellcolor[rgb]{0.6831,0.79,0.9898}\shortstack{\rule{0em}{3ex} -0.0090 \\ 46 / 5 / 61 \\ 0.0502} & \cellcolor[rgb]{0.719,0.812,0.9777}\shortstack{\rule{0em}{3ex} -0.0073 \\ 47 / 5 / 60 \\ 0.0953} & \cellcolor[rgb]{0.8674,0.8644,0.8626}\shortstack{\rule{0em}{3ex} If in bold, then \\ p-value $<$ 0.05} \\[1ex]
\end{tabular}
\begin{tikzpicture}[baseline=(current bounding box.center)]\begin{axis}[hide axis,scale only axis,width=0sp,height=0sp,colorbar horizontal,colorbar style={width=0.25\linewidth,colormap={cm}{rgb255(1)=(83,112,221) rgb255(2)=(220,220,220) rgb255(3)=(209,73,62)},colorbar horizontal,point meta min=-0.03,point meta max=0.03,colorbar/width=1.0em,scaled x ticks=false,xticklabel style={/pgf/number format/fixed,/pgf/number format/precision=3},xlabel={Mean-Difference},}] \addplot[draw=none] {0};\end{axis}\end{tikzpicture}\end{center}
    \caption{Multiple Comparison Matrix for SelF-Rocket with six other methods.}
    \label{fig:MCM}
\end{figure}

\begin{figure*}
        \centering
        \begin{subfigure}[b]{0.475\textwidth}
            \centering
            \includegraphics[width=\textwidth]{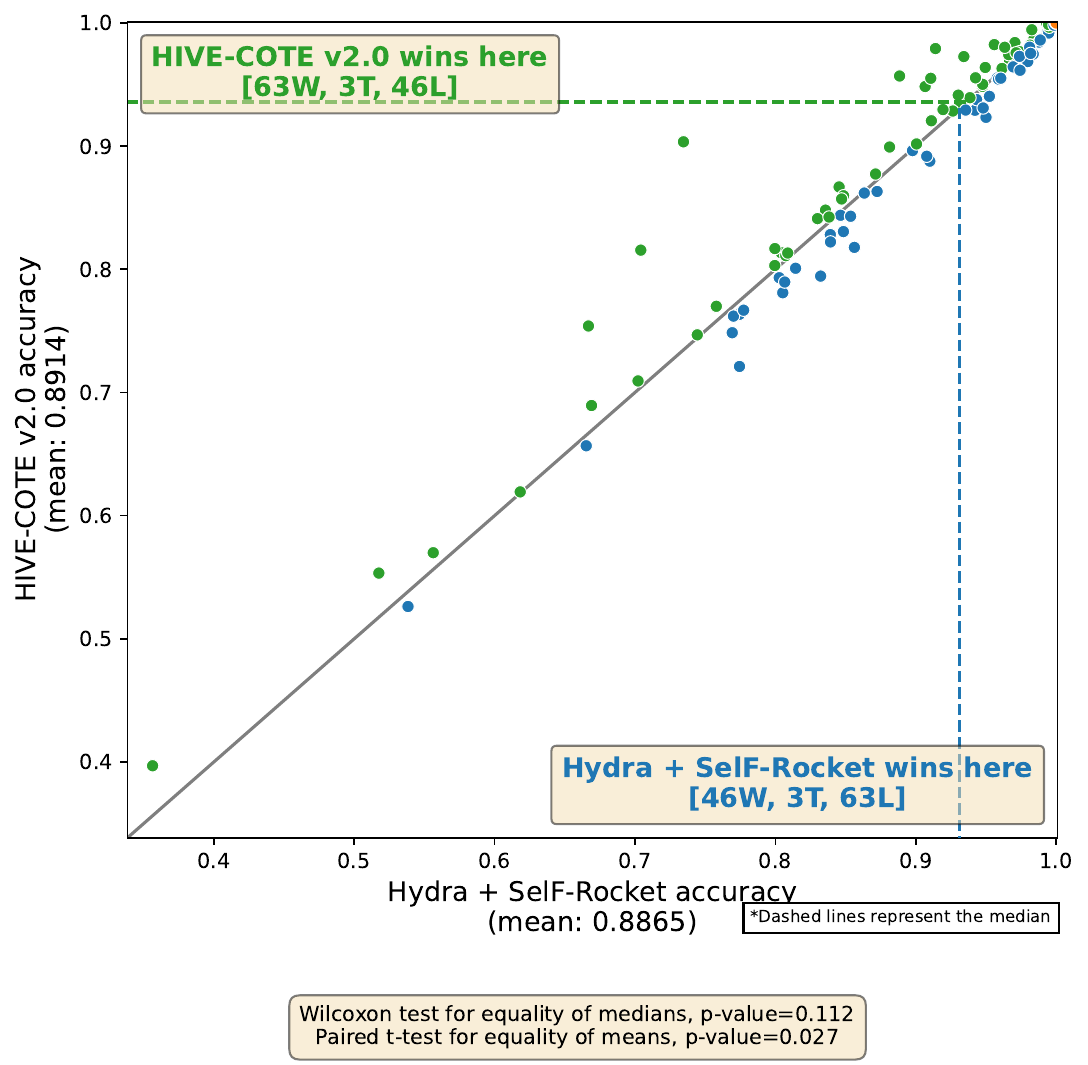}
            \caption[Hydra + SelF-Rocket vs HIVE-COTE v2.0]%
            {{\footnotesize  Hydra + SelF-Rocket vs HIVE-COTE v2.0}}   
            \label{fig:SRHC2}
        \end{subfigure}
        \hfill
        \begin{subfigure}[b]{0.475\textwidth}  
            \centering 
            \includegraphics[width=\textwidth]{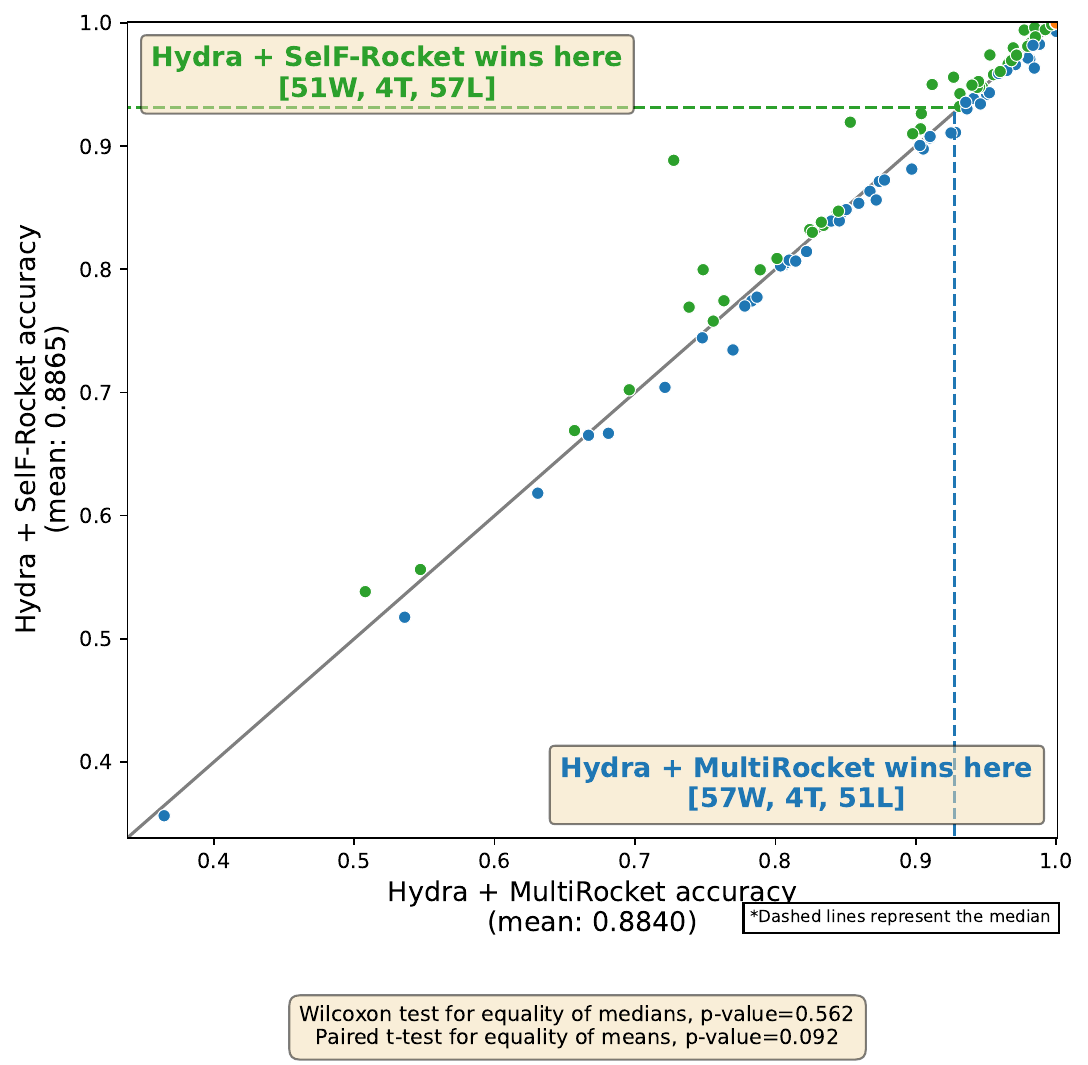}
            \caption[Hydra + SelF-Rocket vs Hydra + MultiRocket]%
            {{\footnotesize Hydra + SelF-Rocket vs Hydra + MultiRocket} }  
            \label{fig:SRHMR}
        \end{subfigure}
        \vskip\baselineskip
        \begin{subfigure}[b]{0.475\textwidth}   
            \centering 
            \includegraphics[width=\textwidth]{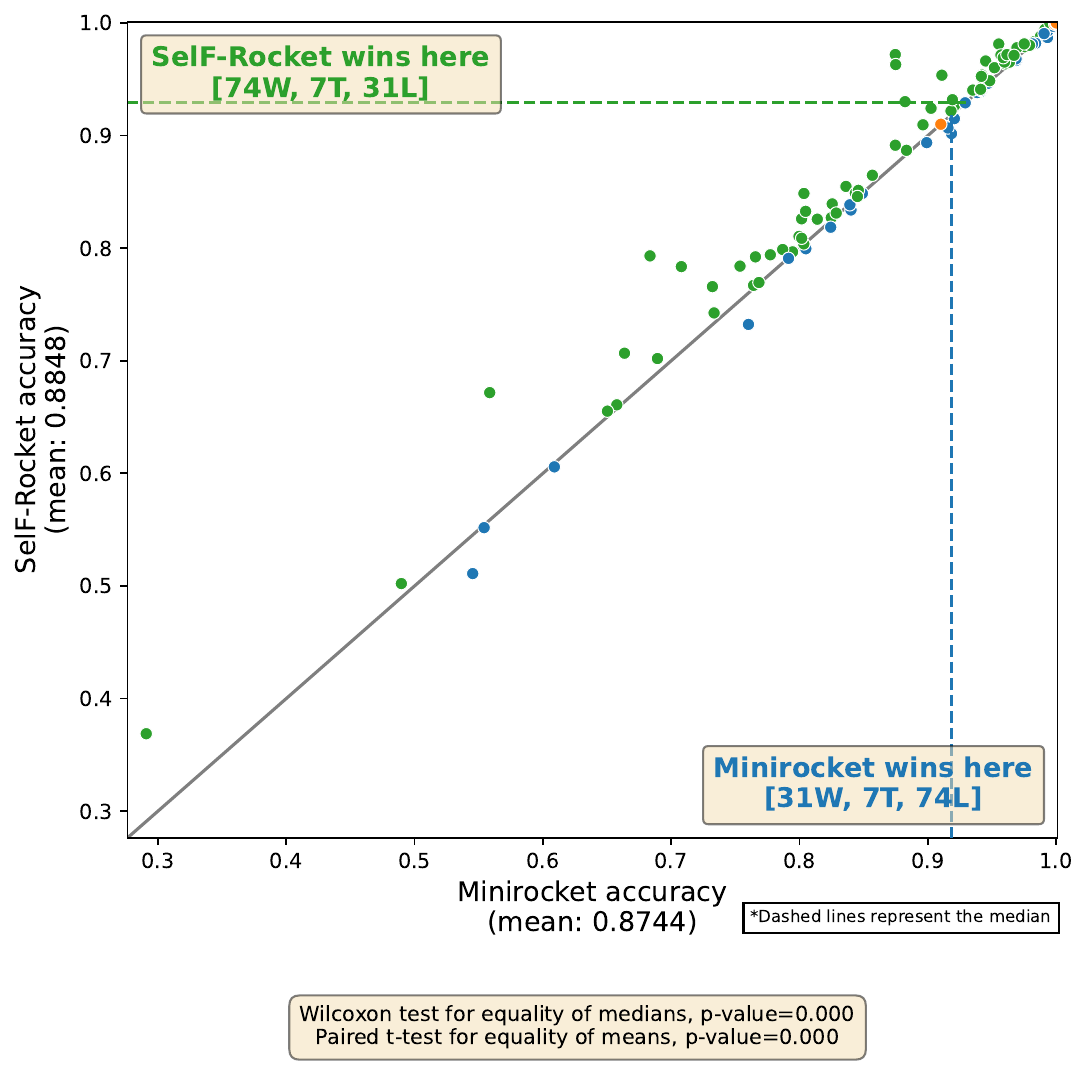}
            \caption[SelF-Rocket vs MiniRocket]%
            {{\footnotesize  SelF-Rocket vs MiniRocket}}    
            \label{fig:SRMR}
        \end{subfigure}
        \hfill
        \begin{subfigure}[b]{0.475\textwidth}   
            \centering 
            \includegraphics[width=\textwidth]{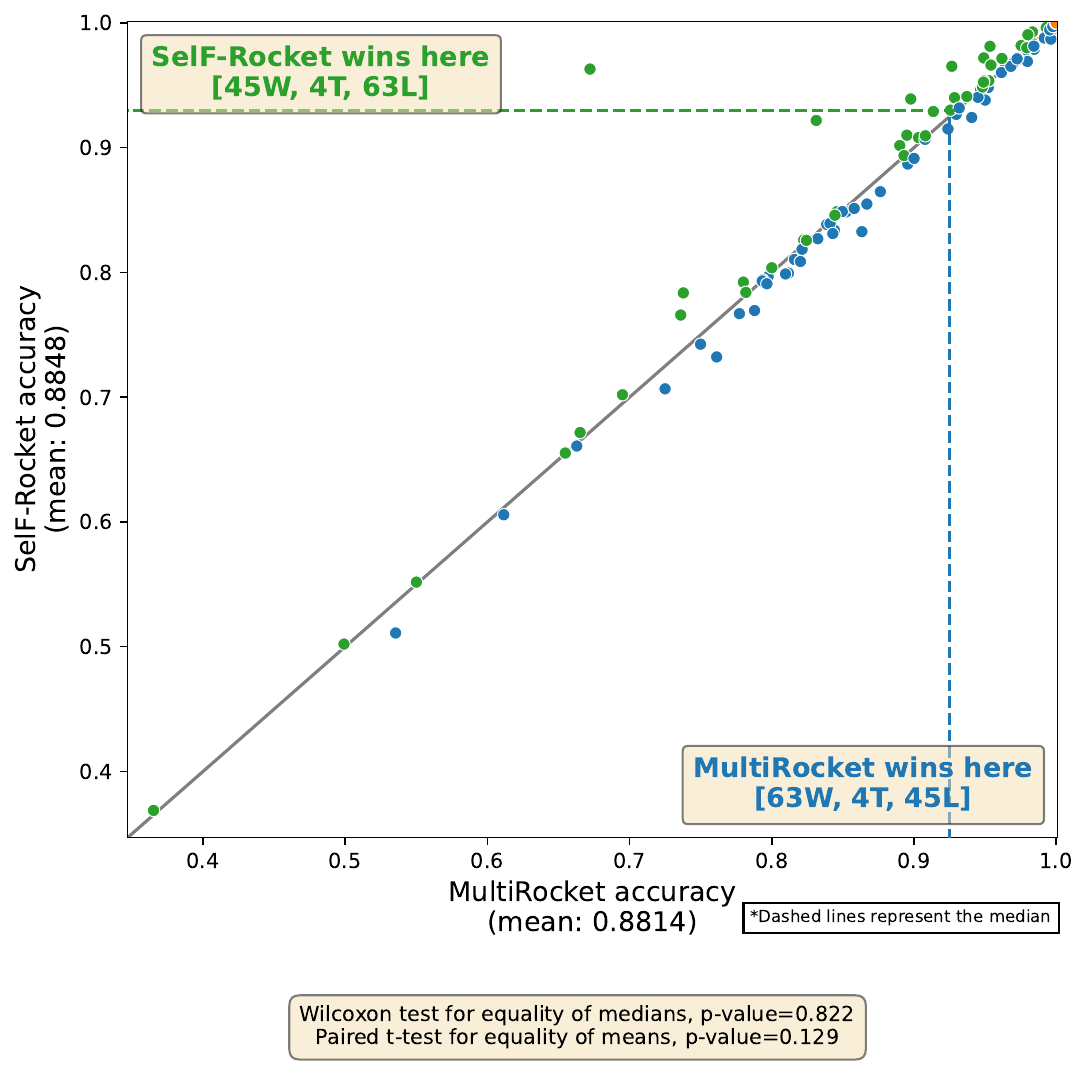}
            \caption[SelF-Rocket vs MultiRocket]%
            {{\footnotesize  SelF-Rocket vs MultiRocket} }  
            \label{fig:SRMLR}
        \end{subfigure}
        \caption[Pairwise accuracy for SelF-Rocket vs State of the Art classifiers]
        {\small Pairwise accuracy for SelF-Rocket vs State of the Art classifiers.}
        \label{fig:SRALL}
    \end{figure*}


The average accuracy rank of these classifiers is displayed in Figure~\ref{fig:ARP}. In terms of performance, Hydra + SelF-Rocket ranks second only to HIVE-COTE v2.0, and outperforms others. SelF-Rocket ranks just after MultiRocket.\newline

Figure~\ref{fig:MCM} depicts the Multiple Comparison Matrix~\cite{ismail2023approach}, i.e. the full pairwise comparison between the best performing methods tested. The mean difference in accuracy between SelF-Rocket and its baseline MiniRocket is approximately 1.04\%. We can also see that SelF-Rocket has a slightly higher accuracy than MultiRocket and Hydra + MultiRocket, despite having fewer wins by pairwise comparison. \newline

Figure~\ref{fig:SRALL} displays the mean accuracy, the pairwise win/draw/loss, and $p$ value for statistical tests between the tested methods over 30 resamples for the 112 selected UCR datasets. Each point represents a dataset; the more a point is distant from the diagonal line, the more one of the two methods performs better than the other. In some cases, Hydra + SelF-Rocket outperforms Hydra + MultiRocket, but in general, both methods offer comparable performance. HIVE-COTE v2.0 has an advantage over Hydra+SelF-Rocket on certain datasets, while SelF-Rocket outperforms MiniRocket and MultiRocket on others. \newline

There is no significant statistical difference by the Wilcoxon signed rank test between Hydra + SelF-Rocket, Hydra + MultiRocket, and HIVE-COTE v2.0. \newline

\subsection{Impact of the Feature Selection Module}

In this section, we first compare the feature selection module with simpler architectures, and examine the effect of vote validation across different dataset sizes.

\begin{figure*}
        \centering
        \begin{subfigure}[t]{0.475\textwidth}
            \centering
            \includegraphics[width=\textwidth]{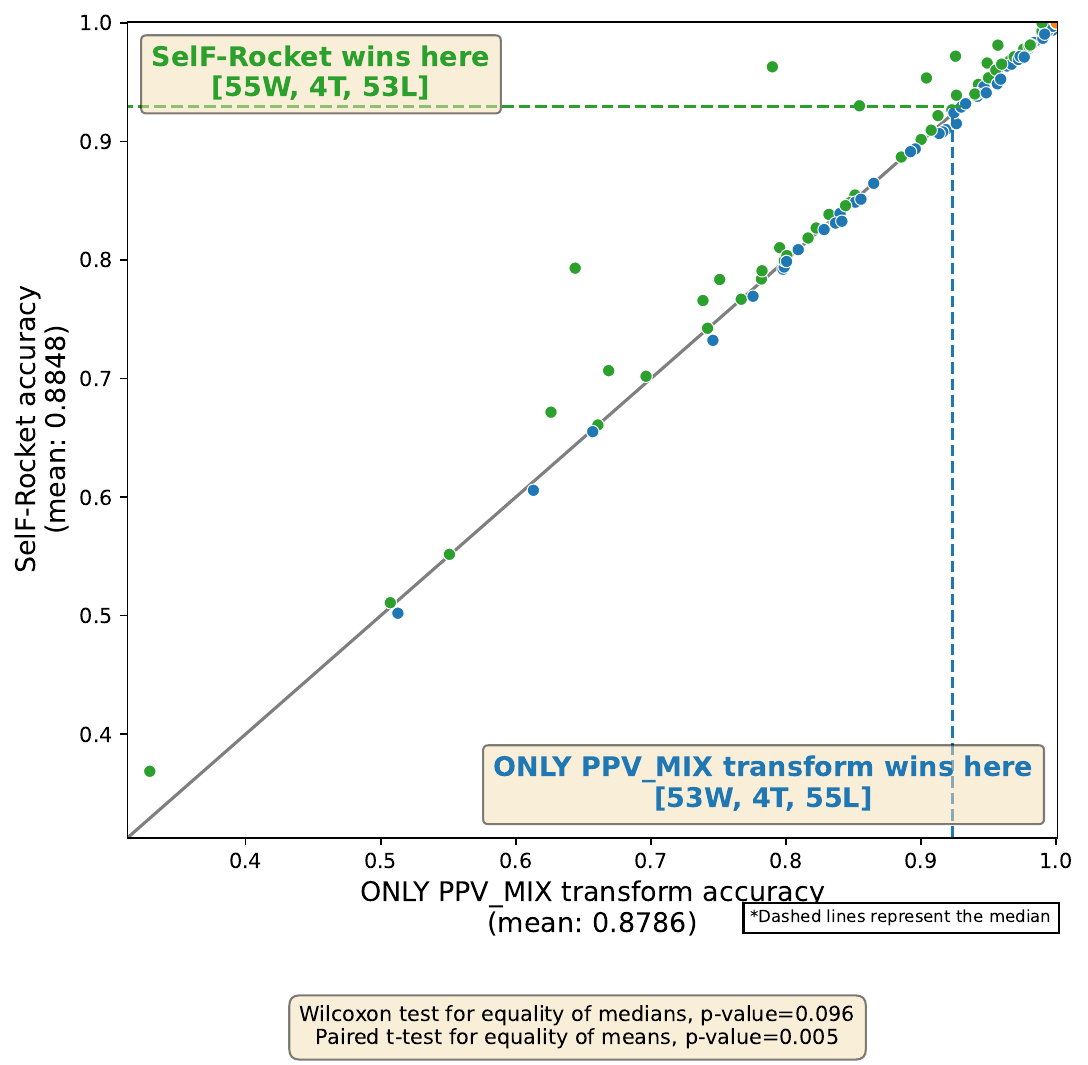}
            \caption[]%
            {{\small SelF-Rocket vs PPV\_MIX}}    
            \label{fig:SRVSPPVMIX}
        \end{subfigure}
        \hfill
        \begin{subfigure}[t]{0.475\textwidth}  
            \centering 
            \includegraphics[width=\textwidth]{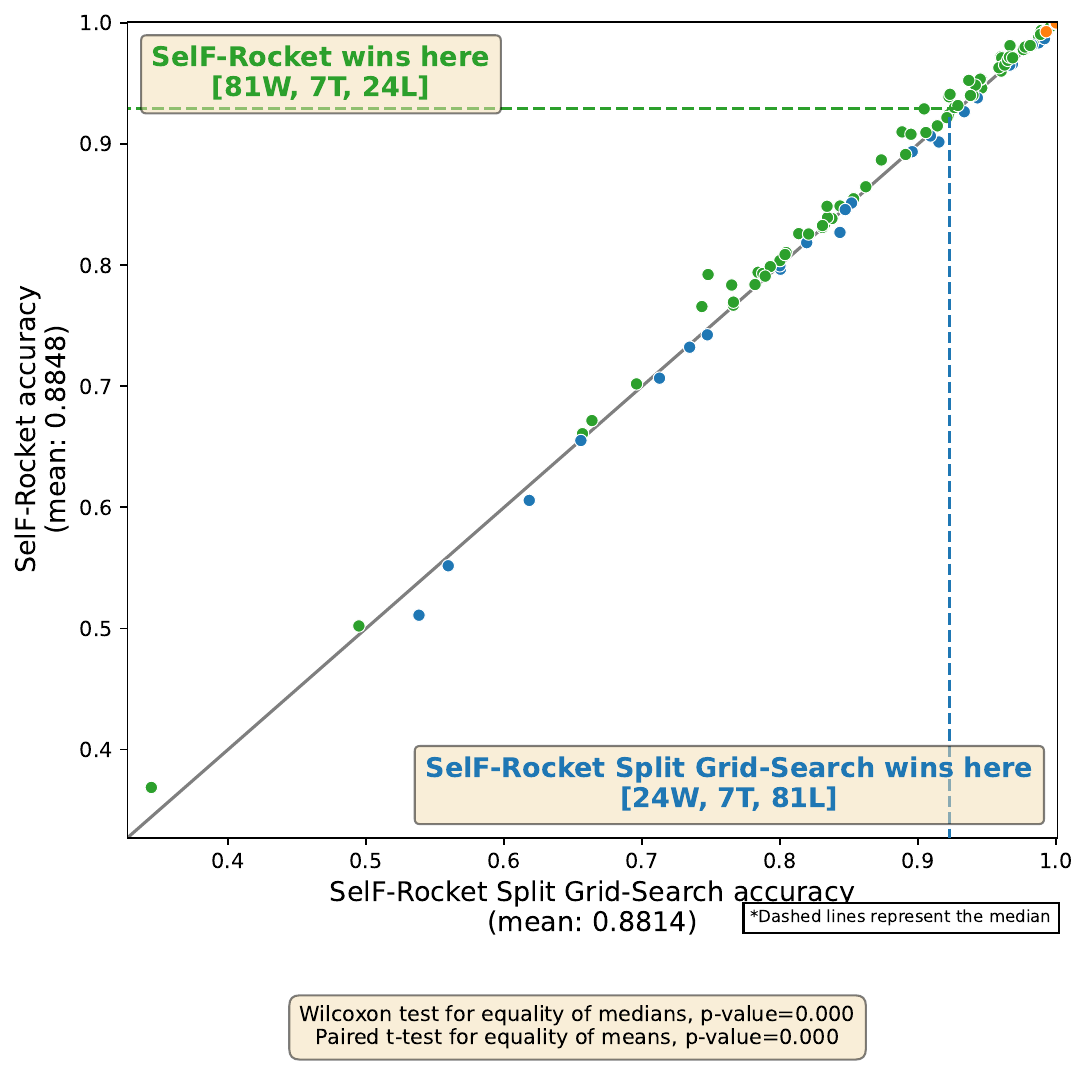}
            \caption[]%
            {{\small SelF-Rocket vs Single Validation Split Grid-Search}}    
            \label{fig:SRVSSVSG}
        \end{subfigure}
        \caption[]
        {\small SelF-Rocket vs simpler controls} 
        \label{fig:SRVSSC}
    \end{figure*}

Figure~\ref{fig:SRVSSC} allows a comparison of this module with systems based on simpler architectures:
Figure~\ref{fig:SRVSPPVMIX} compares SelF-Rocket with the best single transformation namely PPV\_MIX, while Figure~\ref{fig:SRVSSVSG} compares the method with a lightweight feature selection module, in which a single training and validation split is used to evaluate the performance of all possible combinations, and the best IR-PO is selected without vote validation.

In Figure~\ref{fig:SRVSPPVMIX}, SelF-Rocket performs at least as well as PPV\_MIX on all datasets, and substantially better on some of them.
Figure~\ref{fig:SRVSSVSG} shows that, even if SelF-Rocket performs better on most datasets, SelF-Rocket with only one run and a single training–validation split already achieves interesting performance. This strategy could be used for fast training and classification for example.

\begin{figure*}
  \centering
    \begin{subfigure}{\linewidth}
    \centering
    \includegraphics[width=.9\linewidth]{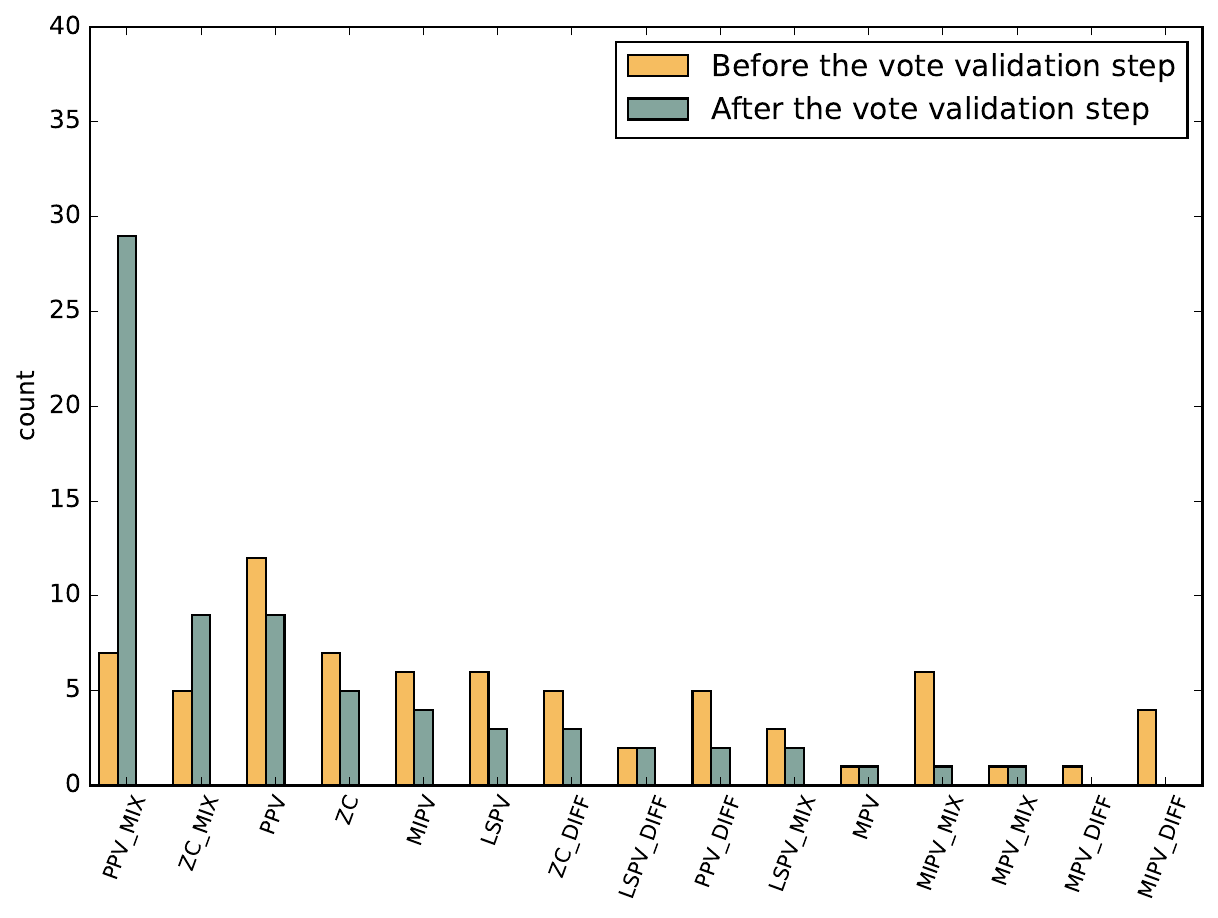}
     \caption[]%
            {{\small  Datasets with more than 100 time series}}    
            \label{fig:DistrIRPOmore100}
\end{subfigure}

  \begin{subfigure}{\linewidth}
    \centering
    \includegraphics[width=.9\linewidth]{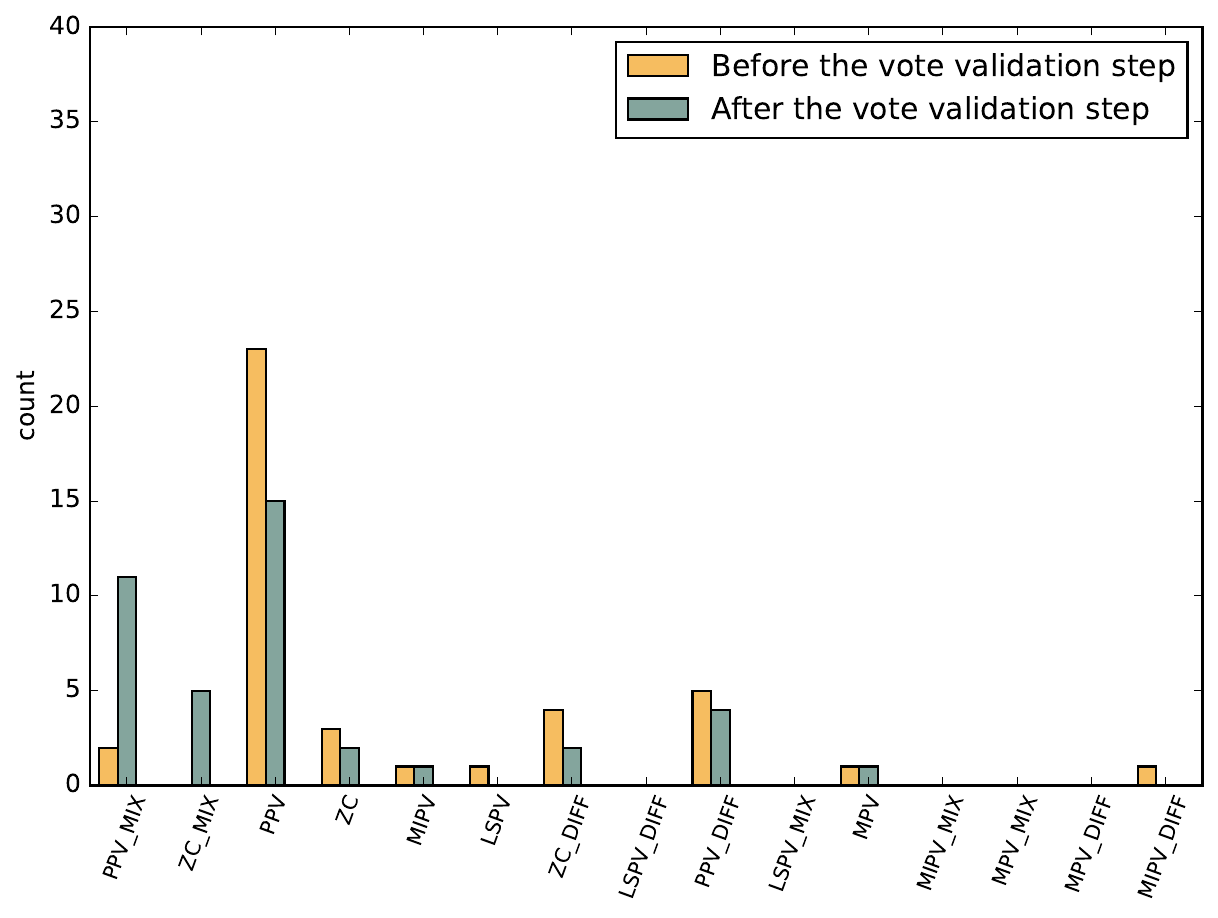}
     \caption[]%
            {{\small  Datasets with less than 100 time series}}    
            \label{fig:DistrIRPOless100}
\end{subfigure}
  \caption[]
        {\small IR-PO combinations selected from the first resample in each of the 112 UCR datasets before and after vote validation. 
        \label{fig:DistrIRPOALL}}  
\end{figure*} 

Table~\ref{tab:MEANDELTAB} presents the difference in average performance between SelF-Rocket with and without vote validation (denoted as $\Delta$ SR/Vote Validation). For small datasets ($\leq 100$), the impact of vote validation is more pronounced compared to datasets with more than 100 instances, where its effect is more limited. In some cases, such as for the ACSF1 dataset, the vote validation module may incorrectly override the original voting results, leading to a negative $\Delta$ SR/Vote Validation. Detailed results are provided in Table~\ref{tab:DELTACONS}.

Figure~\ref{fig:DistrIRPOALL} illustrates the distribution of IR-PO couples under two scenarios: before vote validation and after vote validation, across datasets sizes. We observe a wide variety of selected IR-PO, illustrating the adaptability of the method in choosing different IR-PO depending on the dataset. It can be observed that after vote validation, the proportion of PPV\_MIX and ZC\_MIX increases in both cases. Across the 112 datasets $\times$ 30 resamples runs, we observed 1,389 fallback activations, corresponding to an activation rate of 41.33\%.\newline

\begin{table}[ht]
\centering
\caption{Mean accuracy difference between SelF-Rocket variants (with and without vote validation) and SelF-Rocket–Oracle as a function of dataset size}
\begin{tabular}[t]{|c|c|c|}
\toprule
\multirow{2}{*}{Datasets Size}&\multirow{2}{*}{Mean $\Delta$ SR/Vote Validation}&\multirow{2}{*}{Mean $\Delta$ SR/Oracle}\\
 & &\\
\midrule
$\leq 100$&0.29 \% & -1.09 \%\\
$> 100$ &0.07 \% & -0.46 \% \\
\bottomrule
\end{tabular}
\label{tab:MEANDELTAB}
\end{table}%
    
\subsection{Sensitivity Analysis and Ablative Study}\label{sec:sensitivity}

In this section, we examine the influence of the five key parameters of the SelF-Rocket method on overall accuracy performance:
\begin{itemize}
    \item the Input Representations and Pooling Operators (Section~\ref{sec:irpo}),
    \item the number of folds (Section~\ref{sec:numberoffolds}),
    \item the number of features used to train a mini-classifier (Section~\ref{sec:numberoffeatures}),
    \item the number of runs (Section~\ref{sec:numberofruns}),
    \item and the vote validation parameters (Section~\ref{sec:votingthreshold}). 
\end{itemize}

Following \cite{tan2025proximity,dempster2024quant}, we choose 55 'development' datasets from the 112 UCR datasets to conduct this analysis. We first select the set of datasets that contain at least 5 examples per class and 100 training examples. From that remaining list, 55 datasets were randomly selected. The complete list is available in Table~\ref{fig:TABDEVDAT}.\newline

\subsubsection{Input Representations and Pooling Operators}\label{sec:irpo}

\begin{figure}
    \centering
    \includegraphics[width=0.75\linewidth]{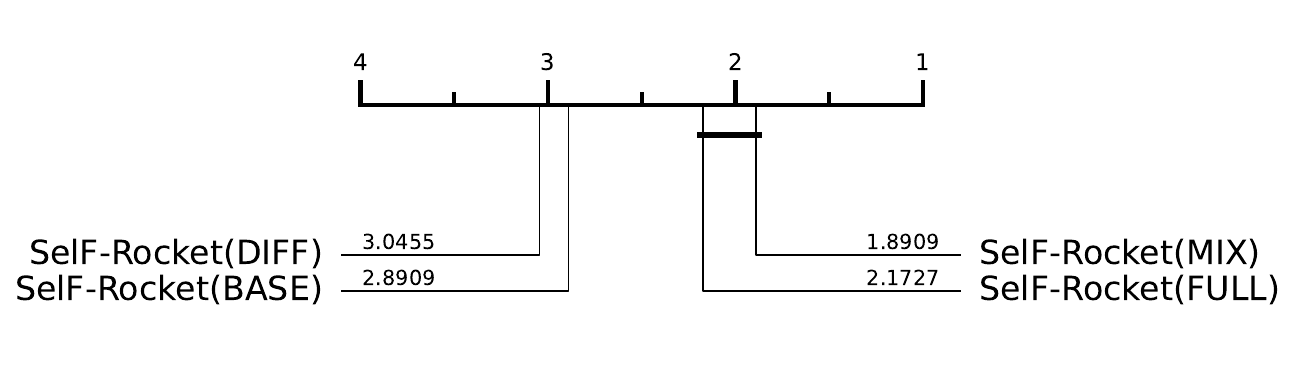}
    \caption{Average mean rank between default SelF-Rocket (FULL) and SelF-Rocket  with only one type of representation.}
    \label{fig:VAREPR}
\end{figure}

\begin{table}[ht]
\centering
\caption{pairwise win/draw/loss between default SelF-Rocket (FULL), and SelF-Rocket with only one type of representation.}
\begin{tabular}[t]{lcccc}
\toprule
&SelF-Rocket&SelF-Rocket&SelF-Rocket&SelF-Rocket\\
&&ONLY BASE&ONLY DIFF&ONLY MIX\\
\midrule
Win&--&11&15&29\\
Draw&--&10&3&6\\
Lose&--&34&37&20\\
Mean accuracy&85.44&84.78&82.79&85.44 \\
\bottomrule
\end{tabular}
\label{fig:VAREPRTAB}
\end{table}%

We evaluated the classification performance of the features generated by 3 sets of input representations: the raw time series ($\text{BASE} \rightarrow \{f(X, \{I \}, p), \forall p \in PO$\}), the first-order difference applied to the time series ($\text{DIFF} \rightarrow \{f(X, \{DIFF \}, p), \forall p \in PO$\}), and a concatenation of features for both ($\text{MIX} \rightarrow \{f(X, \{I, DIFF \}, p), \forall p \in PO$\}). Figure~\ref{fig:VAREPR} illustrates the discrepancy in ranking between the utilization of all three sets of representations and the exclusive use of a single one.
Table~\ref{fig:VAREPRTAB} provides a summary of the pairwise win/draw/loss between default SelF-Rocket (FULL) and SelF-Rocket with only one set of representations. 
The first-order difference (DIFF) is generally the least impactful representation, while the MIX set of representations demonstrates the best overall performance among the three and is similar to the FULL version in terms of performance.\newline

\begin{figure}
    \centering
    \includegraphics[width=0.75\linewidth]{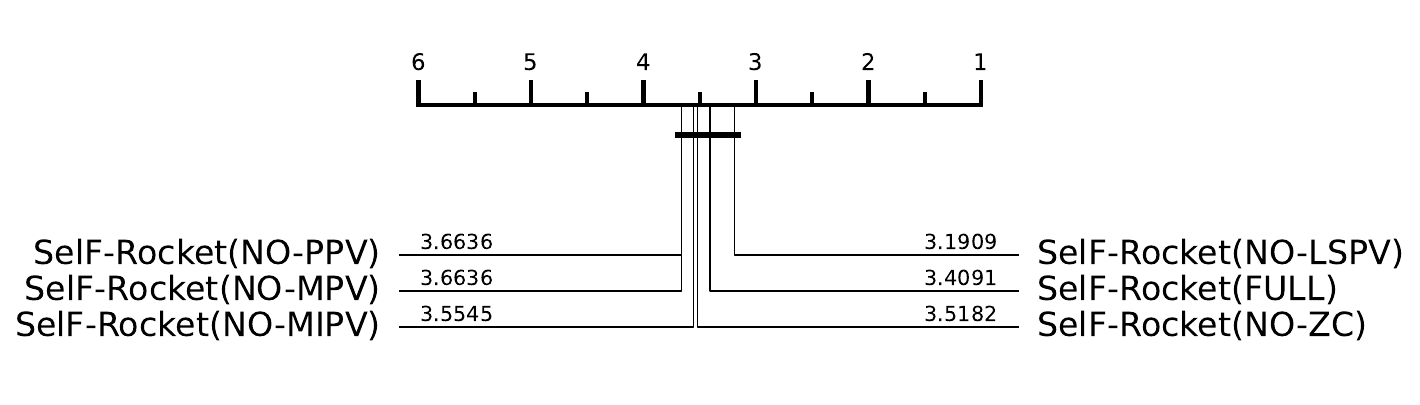}
    \caption{Average mean rank between default SelF-Rocket (FULL) and SelF-Rocket without one type of Pooling Operator.}
    \label{fig:VARPOL}
\end{figure}

\begin{table}[ht]
\centering
\caption{Pairwise win/draw/loss between default SelF-Rocket (FULL), and SelF-Rocket without one type Pooling Operator.}
\begin{tabular}[t]{lcccccc}
\toprule
&\footnotesize{SelF-Rocket}&\footnotesize{SelF-Rocket}&\footnotesize{SelF-Rocket}&\footnotesize{SelF-Rocket}&\footnotesize{SelF-Rocket}&\footnotesize{SelF-Rocket}\\
&&NO PPV&NO ZC&NO MIPV&NO MPV&NO LSPV\\
\midrule
Win&--&18&18&11&11&19\\
Draw&--&10&18&31&26&26\\
Lose&--&27&19&13&18&10\\
Mean acc.&85.44&85.45&85.13&85.21&85.36&85.45 \\
\bottomrule
\end{tabular}
\label{fig:VARPOLTAB}
\end{table}%

Figure~\ref{fig:VARPOL} and Table~\ref{fig:VARPOLTAB} emphasize how not employing one of the five pooling operators in general affects performance. Although no statistically significant difference is observed between the exclusion of any pooling operator and the FULL configuration, removing LSPV results in lower pairwise loss, while removing PPV leads to higher loss\newline

\subsubsection{Number of folds}\label{sec:numberoffolds}

\begin{figure}
    \centering
    \includegraphics[width=0.5\linewidth]{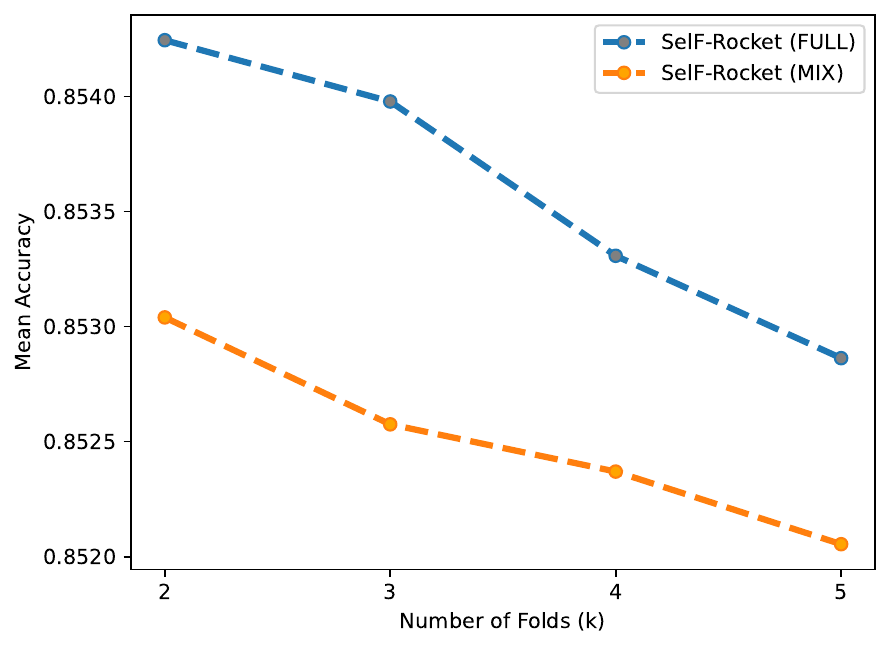}
    \caption{Mean accuracy of SelF-Rocket over 30 resamples across all 55 development datasets, as a function of the number of folds.}
    \label{fig:KFOLDS}
\end{figure}

The SelF-Rocket Feature Selection Module employs a stratified Train\textbackslash Test decomposition of the training set. We have conducted tests using the minimal number of folds possible, namely $k = 2$, up to a fixed value of $k = 5$. Figure~\ref{fig:KFOLDS} illustrates the impact of the number of folds on the classifier mean accuracy. A slight decrease in performance is observed as the number of folds $k$ increases

\subsubsection{Number of features}\label{sec:numberoffeatures}

In its default configuration, MiniRocket only generates 9,996 kernels using PPV as pooling operator, whereas SelF-Rocket employs 9,996 kernels with the base input representation and an additional 9,996 with the first-order difference input representation. 

For each couple (input representation, pooling operator) belonging to $IR \times PO$, SelF-Rocket generates at least 9,996 features. The number of features $f$ tested ranges from 2,500 to 9,996 for each mini-classifier within the Feature Selection Module, $f$ features are randomly selected from the total number of features, which is either 9,996 or 19,992 ($A = \{I,DIFF\}$). 

\begin{figure*}
        \centering
        \begin{subfigure}[b]{0.475\textwidth}
            \centering
            \includegraphics[width=\textwidth]{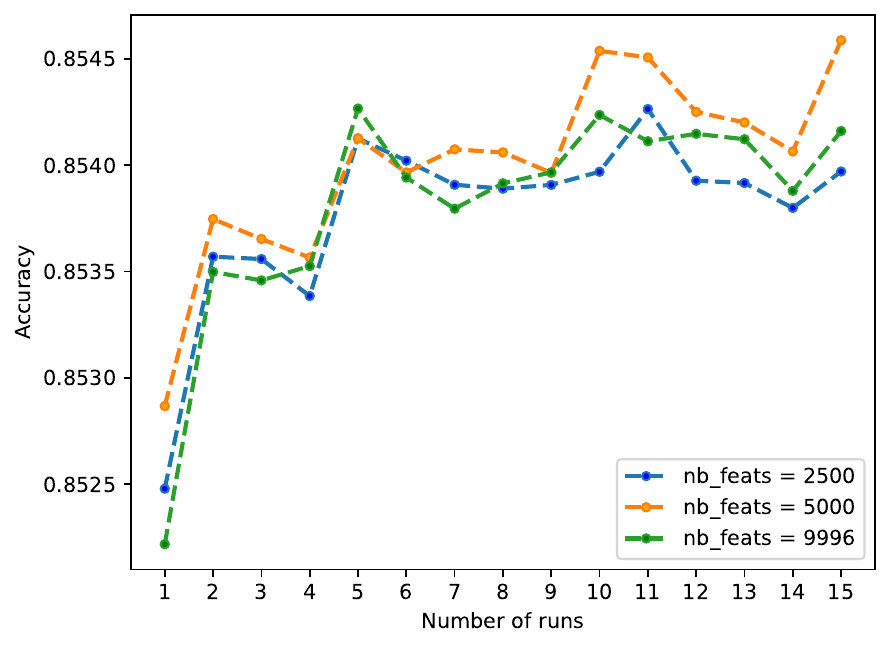}
            \caption[]%
            {{\small SelF-Rocket (FULL)}}    
            \label{fig:NRNFFLA}
        \end{subfigure}
        \hfill
        \begin{subfigure}[b]{0.475\textwidth}  
            \centering 
            \includegraphics[width=\textwidth]{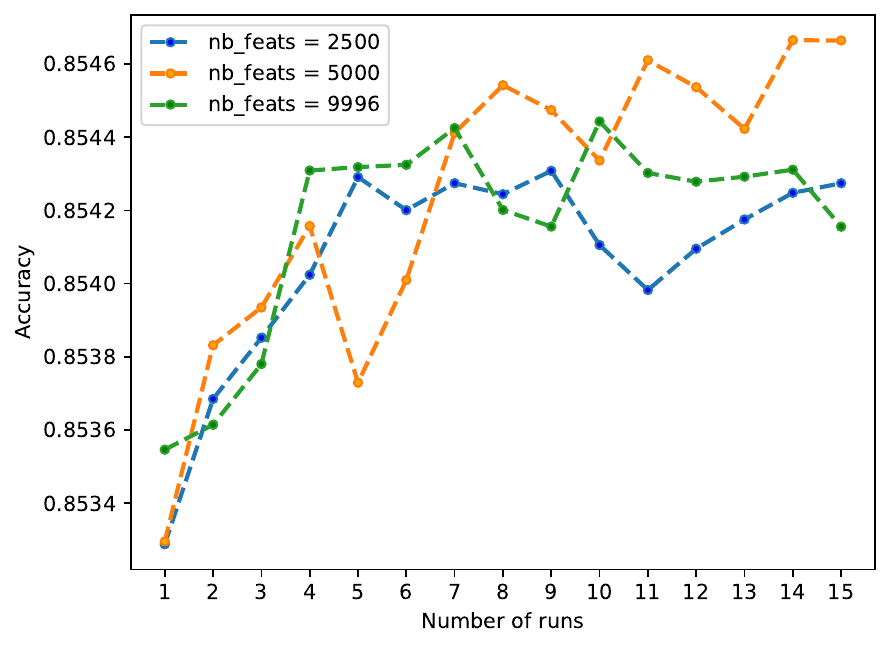}
            \caption[]%
            {{\small SelF-Rocket (MIX)}}    
            \label{fig:NRNFOMA}
        \end{subfigure}
        \caption[]
        {\small Mean accuracy of SelF-Rocket Feature Selection Module over 30 resamples across all 55 development datasets, as a function of the number of runs and selected features.} 
        \label{fig:NRNFA}
    \end{figure*}

As exposed in Figure~\ref{fig:NRNFA}, the number of features has a relatively minor impact on the average accuracy of the final classifier.

\subsubsection{Number of runs}\label{sec:numberofruns}

The total number of decompositions of the initial data set depends on both the number of folds $k$ and the number of runs $nr$. The first one varies the distribution of the number of examples between the training set and the validation set, while the second one allows the original training set to be shuffled and then re-split differently. When $k = 2$, Figure~\ref{fig:NRNFA} shows that the choice of the number of runs $nr$ is relatively more important than the number of features $f$ in order to improve the performance of the Features Selection Module.

\subsubsection{Vote Validation}\label{sec:votingthreshold}

\begin{figure*}
        \centering
        \begin{subfigure}[b]{0.475\textwidth}
            \centering
            \includegraphics[width=\textwidth]{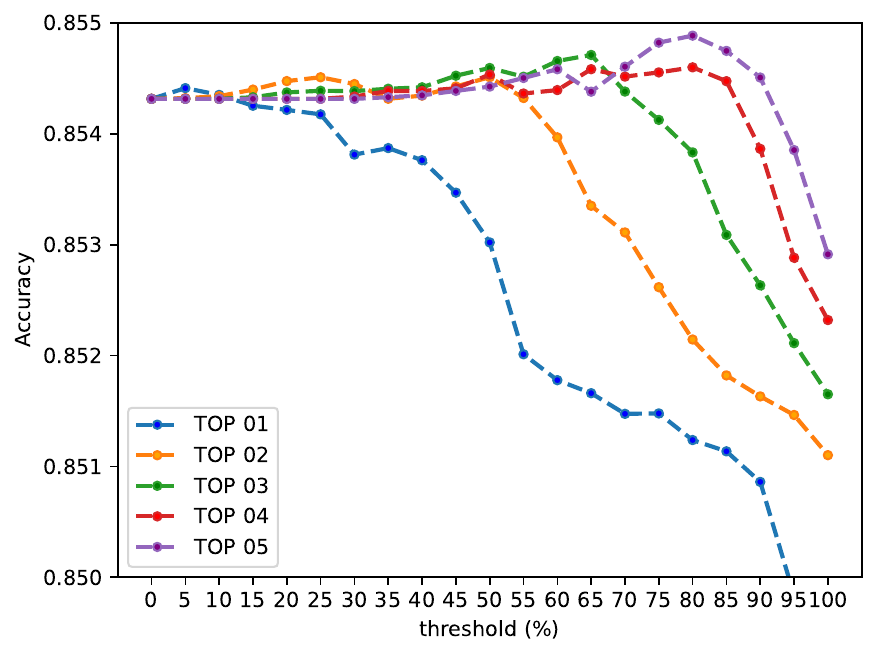}
            \caption[]%
            {{\small  SelF-Rocket (FULL)}}    
            \label{fig:TOPPRCFLA}
        \end{subfigure}
        \hfill
        \begin{subfigure}[b]{0.475\textwidth}  
            \centering 
            \includegraphics[width=\textwidth]{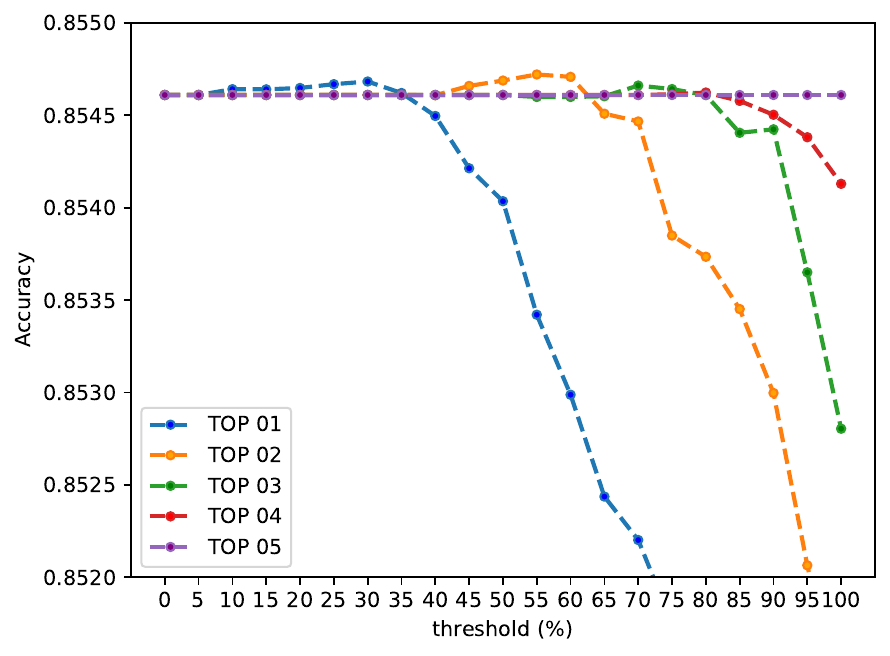}
            \caption[]%
            {{\small SelF-Rocket (MIX)}}    
            \label{fig:TOPPRCOMA}
        \end{subfigure}
        \caption[]
        {\small Mean accuracy of the SelF-Rocket feature selection module over 30 resamples across all 55 development datasets, as a function of the TOP considered and the threshold \textit{thresh} in Algorithm~\ref{alg:VOTVAL}.}
        \label{fig:TOPPRCA}
    \end{figure*}

Once the highest median vote has been carried out, using the result of the vote directly can be risky, especially for small datasets because their mini-classifiers are trained on datasets that are too small when selecting the best set of features. 

To avoid selecting a wrong combination that would generalize poorly, a vote validation system has been put in place by calculating the percentage of voters, who more or less agree with our final choice. If this percentage exceeds a certain value ($\bm{vote_{thresh}}$ in Algorithm~\ref{alg:VOTVAL}) then the vote is maintained, otherwise the vote is replaced by a default choice (PPV\_MIX or ZC\_MIX). Figure~\ref{fig:TOPPRCA} displays the variation of the mean accuracy given a threshold value and a $Top$, e.g. if the $Top = 4$, we calculate the percentage of voters with the final choice in their $Top$ 4 best accuracy. In Figure~\ref{fig:TOPPRCA}, we can see that the higher the top, the higher the threshold \textit{thresh} required to obtain greater accuracy. However, if the threshold is too high, the IR-PO combination selected is replaced too often by the default combination, which gives a lower accuracy.

In this setting, the vote validation system does not yield significant performance improvements, since no dataset with length 
$\leq 100$ is included. Nonetheless, its effect is more substantial on smaller datasets (Table~\ref{tab:MEANDELTAB}).
    
\subsection{Scalability and Compute Time Comparison}
Compared to other ROCKET methods, SelF-Rocket includes a Feature Selection Module between the global feature set and the final classifier. 

\begin{figure*}
        \centering
        \begin{subfigure}[b]{0.475\textwidth}
            \centering
            \includegraphics[width=\textwidth]{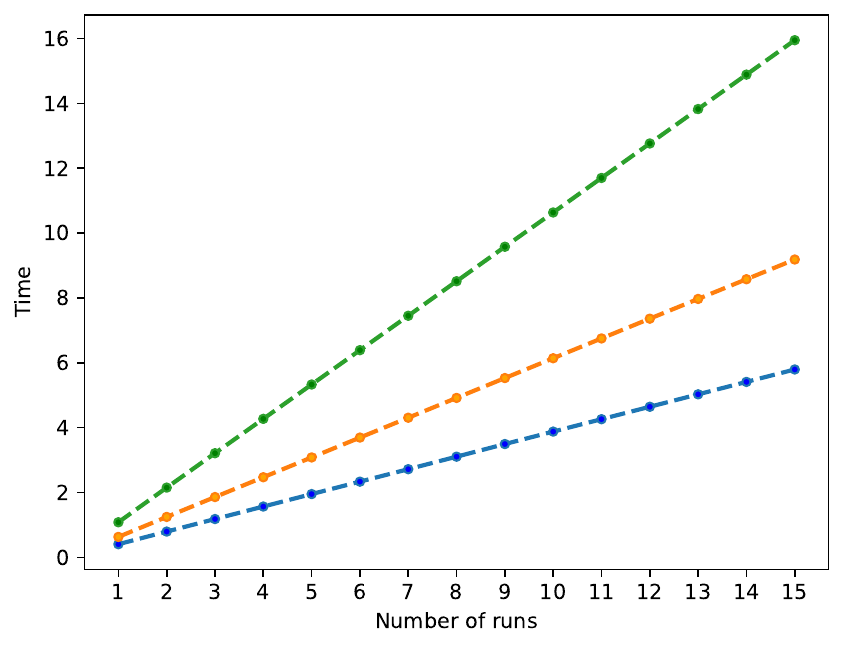}
            \caption[]%
            {{\small SelF-Rocket (FULL)}}    
            \label{fig:CompTimeFSMF}
        \end{subfigure}
        \hfill
        \begin{subfigure}[b]{0.475\textwidth}  
            \centering 
            \includegraphics[width=\textwidth]{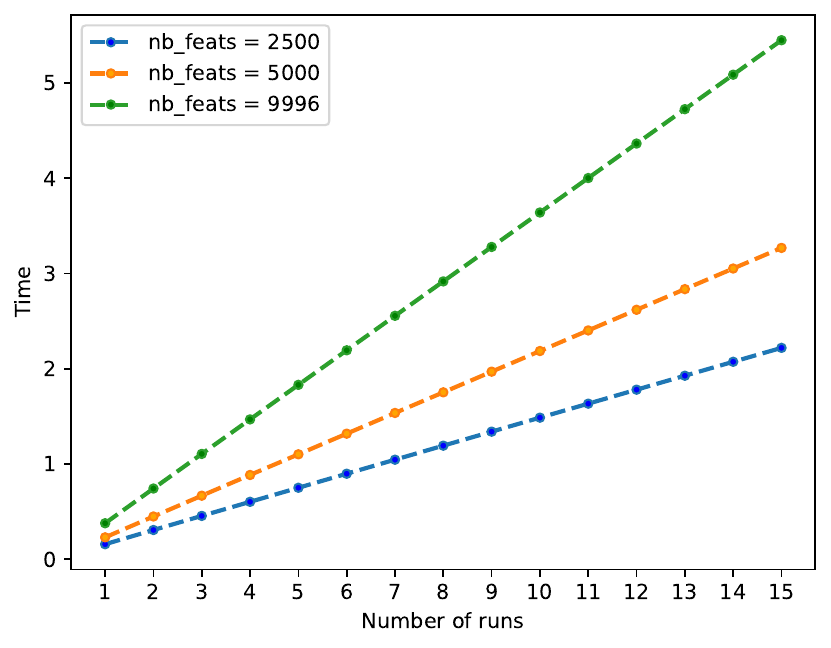}
            \caption[]%
            {{\small SelF-Rocket (MIX)}}    
            \label{fig:CompTimeFSMOM}
        \end{subfigure}
        \caption[]
        {\small Mean computation time of the SelF-Rocket feature selection module, averaged over all 55 development datasets and resamples, as a function of $f$ (number of features) and $nr$ (number of runs).}
        \label{fig:CompTimeFSM}
    \end{figure*}

Figure~\ref{fig:CompTimeFSM} shows the mean computation time for all resamples and all datasets of the Feature Selection Module according to the number of runs and features. The classifier complexity depends on $f$, the number of features and $n$, the number of training examples. The number of runs $nr$ is a multiplier for the number of mini-classifiers, and $f$ impacts the complexity of a single classifier so the two parameters impact generally the compute time. The comparison of the performance of SelF-Rocket (FULL) and SelF-Rocket (MIX) of Section~\ref{sec:irpo} reveals that they are comparable. However, the feature selection phase of SelF-Rocket (MIX) is faster due to a lower number of IR-PO combinations (only 5). With $k=2$, $nr=10$ and $f=2500$, the SelF-Rocket Feature Selection Module takes 1.49 seconds on average for all resamples with the MIX version, compared to 3.88 seconds on average for the FULL version.\newline

\begin{figure*}
        \centering
        \begin{subfigure}[t]{0.475\textwidth}
            \centering
            \includegraphics[width=\textwidth]{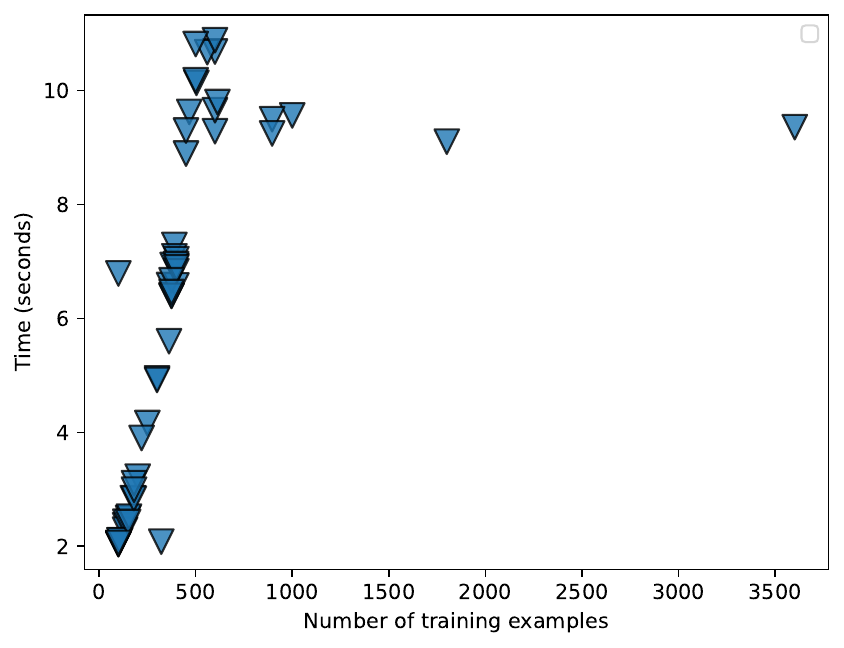}
            \caption[]%
            {{\small  As a function of the number of training examples} }  
            \label{fig:CompTimeNTE}
        \end{subfigure}
        \hfill
        \begin{subfigure}[t]{0.475\textwidth}  
            \centering 
            \includegraphics[width=\textwidth]{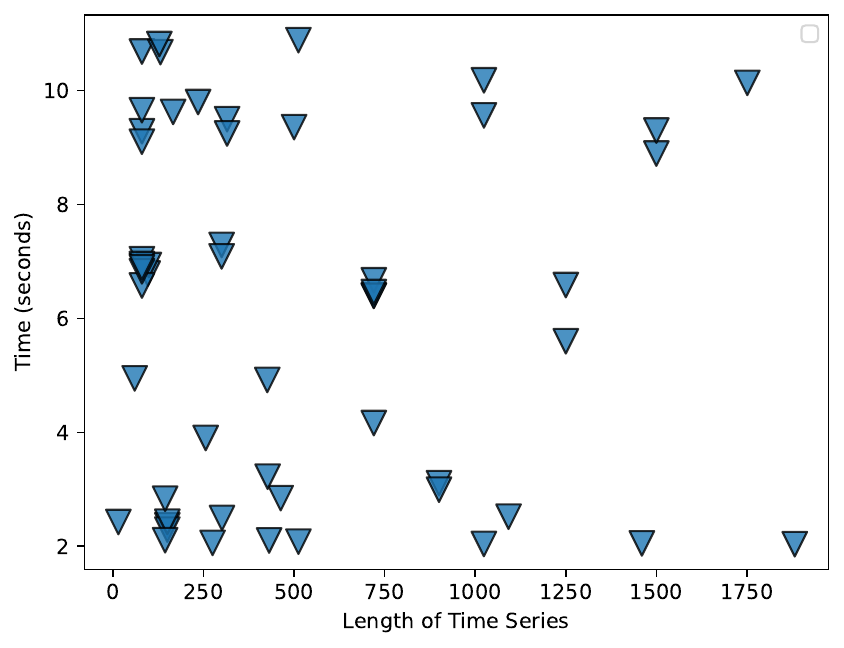}
            \caption[]%
            {{\small As a function of time series length}}   
            \label{fig:CompTimeLTS}
        \end{subfigure}
        \caption[]
        {\small Mean compute time of the SelF-Rocket (FULL, $nr{=}10$, $f{=}5{,}000$) feature selection module, averaged over all 55 development datasets and resamples}
        \label{fig:CompTimeNTELTS}
    \end{figure*}

For each combination of IR and PO, we instantiate $k  \times nr $ mini-classifiers to identify the optimal set of features. A Ridge classifier with a complexity of $\mathcal{O}(n \cdot f^2)$  is employed. In Figure~\ref{fig:CompTimeNTE} given that $f$ is a fixed value, namely 5,000, we can see that the mean time only depends on the number of training examples. We set $mds = 500$ (Algorithm~\ref{alg:FSM}), which explains the plateau observed in the feature selection module's computation time beyond this value. \newline

\begin{table}[htbp]
  \centering
  \caption{Mean training time per resample (in seconds) for SelF-Rocket (MIX, $nr{=}10$, $f{=}2{,}500$), MiniRocket, MultiRocket, Hydra, and Hydra+MultiRocket, including feature creation, feature selection, and ridge training, averaged over the 55 datasets.}
    \begin{tabular}{lcccc}
        \hline
          \multirow{2}{*}{TSC algorithms}              & Features                        &  Features              &  Classifier      & Total             \\
          \\ & Creation & Selection & Train   & Training Time \\ \hline
         MiniRocket     &   5.85         & ---         &  9.70         &  15.55           \\
         MultiRocket    & 34.63        &  ---         & 30.56         &   65.20        \\
         Hydra      &  58.59       &  ---  & 8.03       & 66.62          \\
         SelF-Rocket      & 54.75           & 221.66          & 15.86          & 292.27              \\
          Hydra + MultiRocket      & 85.07           &  ---         & 27.76         & 112.83              \\
         Hydra + SelF-Rocket   & 112.77        &       219.83    &       15.20    &       347.80      \\
        \hline
    \end{tabular}
  \label{tab:grav}
\end{table}

Table~\ref{tab:grav} shows the total computation times over the 55 development datasets, per resample for the ROCKET methods. Among the evaluated methods, MiniRocket remains the fastest overall.
SelF-Rocket features take longer to create than MultiRocket features. This is due to the greater number of kernels that are instantiated. MultiRocket instantiates only 6,216 kernels for each of its input representations (DIFF and BASE). From these, $49,728$ features ($6,216 \text{ kernels} \times 2 \text{ input representations} \times 4 \text{ pooling operators}$) are generated. In comparison, SelF-Rocket generates 9,996 kernels for each of these two representation types for a total of $99,960$ features ($9,996 \text{ kernels} \times 2 \text{ input representations} \times 5 \text{ pooling operators}$).

SelF-Rocket generates approximately nine times more features than MiniRocket during the feature generation phase, but after IR-PO selection it produces either the same number of features or twice as many when the MIX representation is used. The overall computation time can be substantially reduced by limiting the number of kernels generated by SelF-Rocket, decreasing the number of IR-PO (e.g., using only MIX), or reducing the number of runs $nr$.\newline

HYDRA produces $k_g \times g \times d$ kernels~: $k_g$ (number of kernels per group), $g$ (number of groups) and $d$ (maximum possible dilation value), with 2 features extracted per kernel (maximum and minimum responses). For each dataset, 512 ($k_g = 8$ and $g = 64$) kernels per dilation $d$ are created with $2^{d} \leq \text{time series length}$. In most of our cases $d$ will be between 7 and 10, so the total number of features generated will generally be between $512 \times 2 \times 7$ (7,128) and $512 \times 2 \times 10$ (10,240). This explains why HYDRA is the fastest in terms of classifier training time.



\subsection{SelF-Rocket with an oracle}

As shown in Section~\ref{sec:experiments}, selecting the best features can be difficult, particularly in small datasets. Therefore, the implementation of SelF-Rocket presented in this paper is not optimal. Knowing the optimal input representations and pooling operator, what would be the performances of SelF-Rocket? 

\begin{figure*}[ht]
\tiny
\sffamily
  \centering
  \begin{subfigure}{\linewidth}
    \centering
    \includegraphics[width=.8\linewidth]{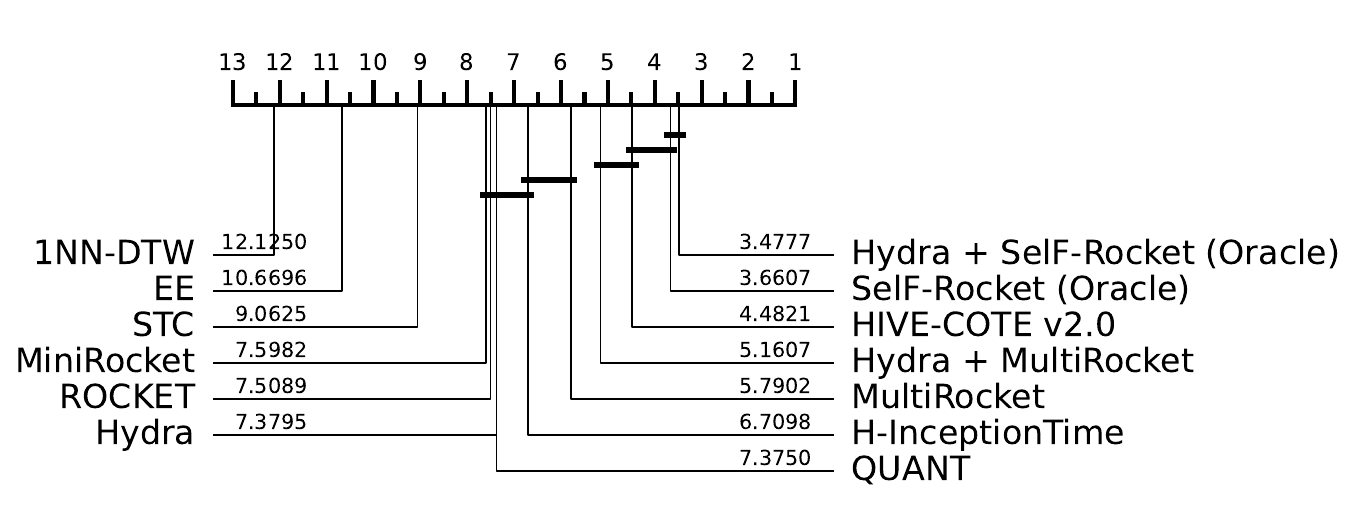}
     \caption[]%
            {{\small  Average mean rank between SelF-Rocket (Oracle), Hydra + SelF-Rocket (Oracle) and others SOTA methods.}}    
            \label{fig:CDORACLE}
  \end{subfigure}
\begin{subfigure}{\linewidth}
\begin{center}
\begin{tabular}{cccccccc}
Mean-Accuracy & \shortstack{H + SelF-R (Oracle) \\ 0.8919} & \shortstack{HC2 \\ 0.8914} & \shortstack{SelF-R (Oracle) \\ 0.8913} & \shortstack{H + MR \\ 0.8840} & \shortstack{MultiR \\ 0.8814} & \shortstack{MiniR \\ 0.8744} \\[1ex]
\shortstack{H + SelF-R (Oracle) \\ 0.8919} & \cellcolor[rgb]{0.8674,0.8644,0.8626}\shortstack{\rule{0em}{3ex} Mean-Difference \\ r$>$c / r=c / r$<$c \\ Wilcoxon p-value} & \bfseries \cellcolor[rgb]{0.8756,0.8602,0.8514}\shortstack{\rule{0em}{3ex} 0.0005 \\ 63 / 6 / 43 \\ 0.0169} & \cellcolor[rgb]{0.8796,0.8582,0.8458}\shortstack{\rule{0em}{3ex} 0.0006 \\ 56 / 8 / 48 \\ 0.1742} & \bfseries \cellcolor[rgb]{0.9697,0.6905,0.5751}\shortstack{\rule{0em}{3ex} 0.0079 \\ 74 / 8 / 30 \\  $\leq$ 1e-04} & \bfseries \cellcolor[rgb]{0.9583,0.6043,0.4833}\shortstack{\rule{0em}{3ex} 0.0104 \\ 81 / 8 / 23 \\  $\leq$ 1e-04} & \bfseries \cellcolor[rgb]{0.8204,0.2868,0.2452}\shortstack{\rule{0em}{3ex} 0.0175 \\ 99 / 5 / 8 \\  $\leq$ 1e-04} \\[1ex]
\shortstack{HC2 \\ 0.8914} & \bfseries \cellcolor[rgb]{0.8554,0.8638,0.8766}\shortstack{\rule{0em}{3ex} -0.0005 \\ 43 / 6 / 63 \\ 0.0169} & \cellcolor[rgb]{0.8674,0.8644,0.8626}\shortstack{\rule{0em}{3ex} -} & \cellcolor[rgb]{0.8674,0.8644,0.8626}\shortstack{\rule{0em}{3ex} 0.0001 \\ 47 / 6 / 59 \\ 0.2821} & \cellcolor[rgb]{0.9692,0.7058,0.5937}\shortstack{\rule{0em}{3ex} 0.0074 \\ 59 / 7 / 46 \\ 0.1046} & \bfseries \cellcolor[rgb]{0.9616,0.6222,0.5016}\shortstack{\rule{0em}{3ex} 0.0100 \\ 67 / 6 / 39 \\ 0.0051} & \bfseries \cellcolor[rgb]{0.835,0.3136,0.2598}\shortstack{\rule{0em}{3ex} 0.0170 \\ 84 / 5 / 23 \\  $\leq$ 1e-04} \\[1ex]
\shortstack{SelF-R (Oracle) \\ 0.8913} & \cellcolor[rgb]{0.8514,0.8631,0.8811}\shortstack{\rule{0em}{3ex} -0.0006 \\ 48 / 8 / 56 \\ 0.1742} & \cellcolor[rgb]{0.8634,0.8651,0.8676}\shortstack{\rule{0em}{3ex} -0.0001 \\ 59 / 6 / 47 \\ 0.2821} & \cellcolor[rgb]{0.8674,0.8644,0.8626}\shortstack{\rule{0em}{3ex} -} & \bfseries \cellcolor[rgb]{0.9689,0.7108,0.5999}\shortstack{\rule{0em}{3ex} 0.0073 \\ 72 / 7 / 33 \\  $\leq$ 1e-04} & \bfseries \cellcolor[rgb]{0.9627,0.6282,0.5076}\shortstack{\rule{0em}{3ex} 0.0098 \\ 80 / 7 / 25 \\  $\leq$ 1e-04} & \bfseries \cellcolor[rgb]{0.8394,0.3219,0.2649}\shortstack{\rule{0em}{3ex} 0.0169 \\ 106 / 5 / 1 \\  $\leq$ 1e-04} \\[1ex]
\shortstack{H + MR \\ 0.8840} & \bfseries \cellcolor[rgb]{0.6355,0.7567,0.9983}\shortstack{\rule{0em}{3ex} -0.0079 \\ 30 / 8 / 74 \\  $\leq$ 1e-04} & \cellcolor[rgb]{0.6514,0.7681,0.9959}\shortstack{\rule{0em}{3ex} -0.0074 \\ 46 / 7 / 59 \\ 0.1046} & \bfseries \cellcolor[rgb]{0.6567,0.7718,0.9949}\shortstack{\rule{0em}{3ex} -0.0073 \\ 33 / 7 / 72 \\  $\leq$ 1e-04} & \cellcolor[rgb]{0.8674,0.8644,0.8626}\shortstack{\rule{0em}{3ex} -} & \bfseries \cellcolor[rgb]{0.9227,0.8286,0.7771}\shortstack{\rule{0em}{3ex} 0.0026 \\ 68 / 9 / 35 \\ 0.0003} & \bfseries \cellcolor[rgb]{0.9638,0.6342,0.5137}\shortstack{\rule{0em}{3ex} 0.0096 \\ 80 / 5 / 27 \\  $\leq$ 1e-04} \\[1ex]
\shortstack{MultiR \\ 0.8814} & \bfseries \cellcolor[rgb]{0.5543,0.6901,0.9955}\shortstack{\rule{0em}{3ex} -0.0104 \\ 23 / 8 / 81 \\  $\leq$ 1e-04} & \bfseries \cellcolor[rgb]{0.5706,0.7041,0.9972}\shortstack{\rule{0em}{3ex} -0.0100 \\ 39 / 6 / 67 \\ 0.0051} & \bfseries \cellcolor[rgb]{0.5761,0.7088,0.9978}\shortstack{\rule{0em}{3ex} -0.0098 \\ 25 / 7 / 80 \\  $\leq$ 1e-04} & \bfseries \cellcolor[rgb]{0.8006,0.8504,0.93}\shortstack{\rule{0em}{3ex} -0.0026 \\ 35 / 9 / 68 \\ 0.0003} & \cellcolor[rgb]{0.8674,0.8644,0.8626}\shortstack{\rule{0em}{3ex} -} & \bfseries \cellcolor[rgb]{0.9682,0.7208,0.6123}\shortstack{\rule{0em}{3ex} 0.0070 \\ 80 / 5 / 27 \\  $\leq$ 1e-04} \\[1ex]
\shortstack{MiniR \\ 0.8744} & \bfseries \cellcolor[rgb]{0.3286,0.4397,0.8696}\shortstack{\rule{0em}{3ex} -0.0175 \\ 8 / 5 / 99 \\  $\leq$ 1e-04} & \bfseries \cellcolor[rgb]{0.3433,0.4594,0.8841}\shortstack{\rule{0em}{3ex} -0.0170 \\ 23 / 5 / 84 \\  $\leq$ 1e-04} & \bfseries \cellcolor[rgb]{0.3483,0.4657,0.8883}\shortstack{\rule{0em}{3ex} -0.0169 \\ 1 / 5 / 106 \\  $\leq$ 1e-04} & \bfseries \cellcolor[rgb]{0.5815,0.7135,0.9983}\shortstack{\rule{0em}{3ex} -0.0096 \\ 27 / 5 / 80 \\  $\leq$ 1e-04} & \bfseries \cellcolor[rgb]{0.6673,0.7792,0.993}\shortstack{\rule{0em}{3ex} -0.0070 \\ 27 / 5 / 80 \\  $\leq$ 1e-04} & \cellcolor[rgb]{0.8674,0.8644,0.8626}\shortstack{\rule{0em}{3ex} If in bold, then \\ p-value $<$ 0.05} \\[1ex]
\end{tabular}
\begin{tikzpicture}[baseline=(current bounding box.center)]\begin{axis}[hide axis,scale only axis,width=1pt,colorbar right,colorbar style={height=0.25\linewidth,colormap={cm}{rgb255(1)=(83,112,221) rgb255(2)=(220,220,220) rgb255(3)=(209,73,62)},colorbar horizontal,point meta min=-0.02,point meta max=0.02,colorbar/width=1.0em,scaled y ticks=false,ylabel style={rotate=180},yticklabel style={/pgf/number format/fixed,/pgf/number format/precision=3},ylabel={Mean-Difference},}] \addplot[draw=none] {0};\end{axis}\end{tikzpicture}\end{center}
    \caption[]%
            {{\small Multiple Comparison Matrix for SelF-Rocket (Oracle), Hydra + SelF-Rocket (Oracle)  with other methods.}}   
            \label{fig:HMORACLE}
  \end{subfigure}
  \caption[]
        {\small Performance of SelF-Rocket (Oracle) and Hydra + SelF-Rocket (Oracle) over 30 resamples across all the 112 UCR datasets. 
        \label{fig:SRORACLE}}  
\end{figure*}

Figure~\ref{fig:CDORACLE} illustrates the average mean accuracy ranking for the various classifiers previously tested, while Figure~\ref{fig:HMORACLE} presents the MCM of the Oracle version of SelF-Rocket and Hydra combined with SelF-Rocket, compared to HIVE-COTE v2.0 and other ROCKET methods. These results indicate the maximum potential performance of an algorithm based on SelF-Rocket. The question remains whether such performance is attainable, or if we can approach it with an improved Feature Selection Module.

\section{Conclusion and prospects}\label{sec:conclusionandprospects}

This article presents a novel algorithm called SelF-Rocket, which builds upon MiniRocket by incorporating a feature selection stage. 
The key idea behind SelF-Rocket is its dynamic selection of the best combination of input representations and pooling operators during the training phase. Experiments conducted on the UCR datasets show that SelF-Rocket, on average, selects a more effective combination of Input Representation and Pooling Operator compared to MiniRocket, and succeeds to achieve state-of-the-art performance. SelF-Rocket delivers performance comparable to MultiRocket (similarly, Hydra combined with SelF-Rocket matches the performance of Hydra combined with MultiRocket). However, SelF-Rocket offers the benefit of quicker classification prediction times due to its utilization of fewer features (9,996 or 19,992 for SelF-Rocket vs 50,000 for MultiRocket).\newline 



We also provide an estimate of the idealized performance that could be achieved if the feature selection module were optimal in selecting the input representation and pooling operator, illustrating the potential of the method while leaving open the question of whether such performance is attainable in practice.

The version of SelF-Rocket evaluated in this study employs three sets of input representations, five pooling operators, and a simple wrapper-based feature selection method. Potential improvements to this implementation could include:
\begin{itemize}
    \item Incorporating additional input representations, such as second-order differences, Fourier transforms, and Hilbert transforms, among others.
    \item Improving the vote validation systems could potentially bring performance closer to the optimal levels achieved by the oracle.
    \item Using a filter-based feature selection method (based on $\chi^2$ test for instance) instead of Stratified $k$-Fold to speed-up the feature selection stage.
\end{itemize}

\section*{Acknowledgments}
This research project, supported and financed by the French ANR (Agence Nationale pour la Recherche), is part of the Labcom (Laboratoire Commun) MYEL (MobilitY and Reliability of Electrical chain Lab) involving LSEE, LGI2A and CRITTM2A (ANR-22-LCV2-0001 MYEL).

The authors would like to thank Prof. Eamonn Keogh and all those who have contributed, and continue to contribute, to the maintenance of the University of California Riverside (UCR) TSC benchmark datasets and to open source implementations of the algorithms used in this article:
\begin{itemize}
    \item The original implementations of MiniRocket\footnote{\url{https://github.com/angus924/minirocket}}, MultiRocket\footnote{\url{https://github.com/ChangWeiTan/MultiRocket}} and Hydra\footnote{\url{https://github.com/angus924/hydra}} were used as baseline.
    \item Critical Difference diagrams were generated using aeon toolkit~\cite{aeon24jmlr}.
    \item Multiple Comparison Matrices were generated using the original implementation\footnote{\url{https://github.com/MSD-IRIMAS/Multi_Comparison_Matrix}} of~\cite{ismailfawaz2023approachmultiplecomparisonbenchmark}. 
    \item The results provided by tsml-eval\footnote{\url{https://github.com/time-series-machine-learning/tsml-eval}} were used to benchmark SelF-Rocket.
\end{itemize}

\begin{appendices}

\section{Additional Figures \& Tables}\label{sec:A1}

\begin{figure}[!h]
    \centering
    \includegraphics[width=\linewidth]{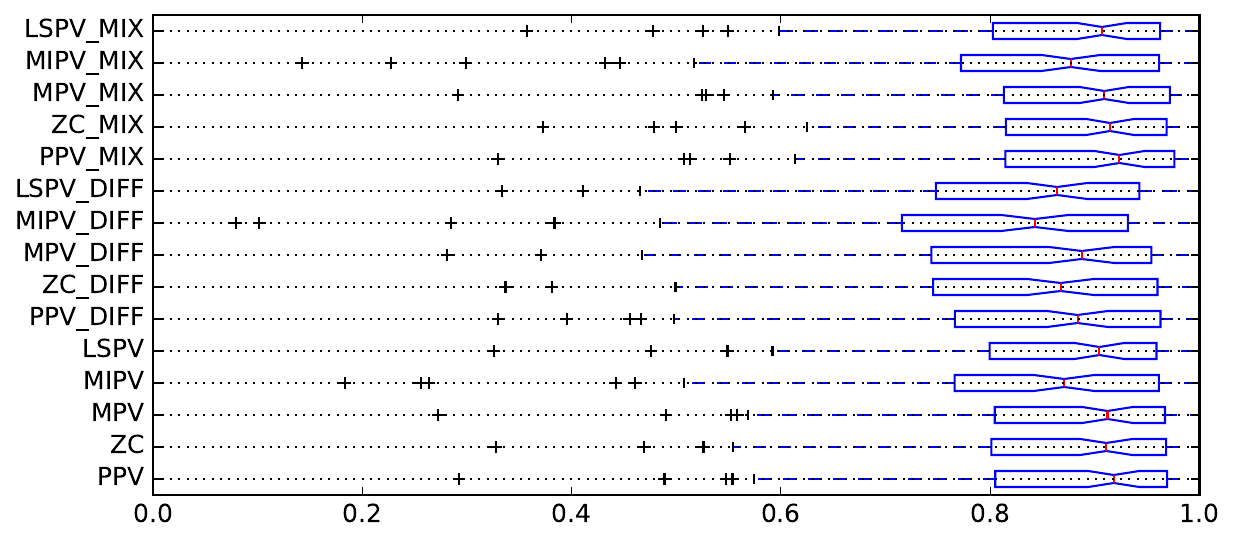}
    \caption{Distribution of accuracy for the 15 possible transformations over 30 resamples on the 112 selected UCR datasets}
    \label{fig:Distrib_PO_Detailed}
\end{figure}

\begin{table}[!ht]
    \centering
    \caption{List of 55 development datasets used in the sensibility analysis.}
    \begin{tabular}{|c||c||c|}
        \hline
        \multicolumn{3}{|c|}{Development datasets} \\
        \hline
        ACSF1 & ChlorineConcentration & Computers \\
        CricketX & CricketY &  DistalPhalanxOutlineAgeGroup \\ 
        DistalPhalanxOutlineCorrect & DistalPhalanxTW & Earthquakes \\
        ECG200 & EOGHorizontalSignal & EOGVerticalSignal \\
        EthanolLevel & FaceAll & Fish \\
        FordA & FreezerRegularTrain & GunPointAgeSpan \\
        GunPointMaleVersusFemale & GunPointOldVersusYoung & Ham \\
        Haptics & InlineSkate & InsectWingbeatSound \\
        LargeKitchenAppliances & MedicalImages & MiddlePhalanxOutlineAgeGroup \\
        MiddlePhalanxOutlineCorrect & MiddlePhalanxTW & MixedShapesRegularTrain \\ MixedShapesSmallTrain & OSULeaf & PhalangesOutlinesCorrect \\
        Plane & PowerCons & ProximalPhalanxOutlineAgeGroup \\
        ProximalPhalanxOutlineCorrect & ProximalPhalanxTW & RefrigerationDevices \\
        ScreenType  & SemgHandMovementCh2 & SemgHandSubjectCh2 \\
        ShapesAll & SmallKitchenAppliances & SmoothSubspace \\
        StarLightCurves & Strawberry & SwedishLeaf \\
        SyntheticControl & Trace & UWaveGestureLibraryX \\ 
        UWaveGestureLibraryZ & Worms & WormsTwoClass  \\ 
        Yoga & & \\ \hline
    \end{tabular}
    \label{fig:TABDEVDAT}
\end{table}

\begin{figure}
    \centering
    \includegraphics[width=1\linewidth]{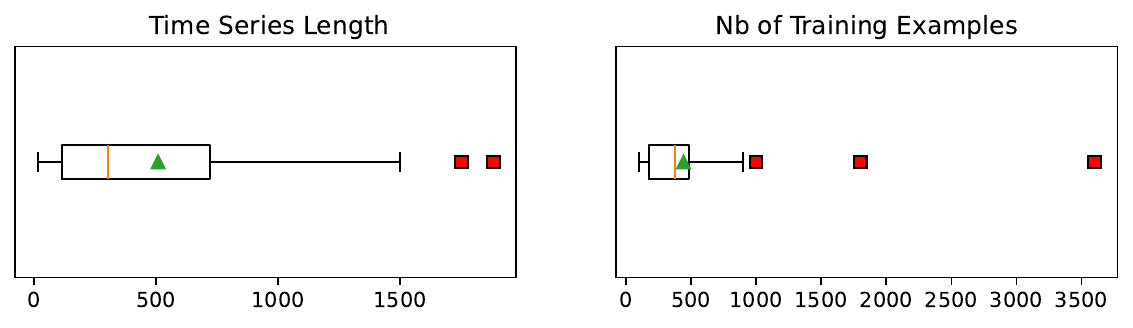}
    \caption{Distribution of the time series length of the development datasets (left), distribution of the number of training examples of the development datasets (right)}
    \label{fig:Distriblenbtrex}
\end{figure}

\begin{figure}
    \centering
    \includegraphics[width=0.6\linewidth]{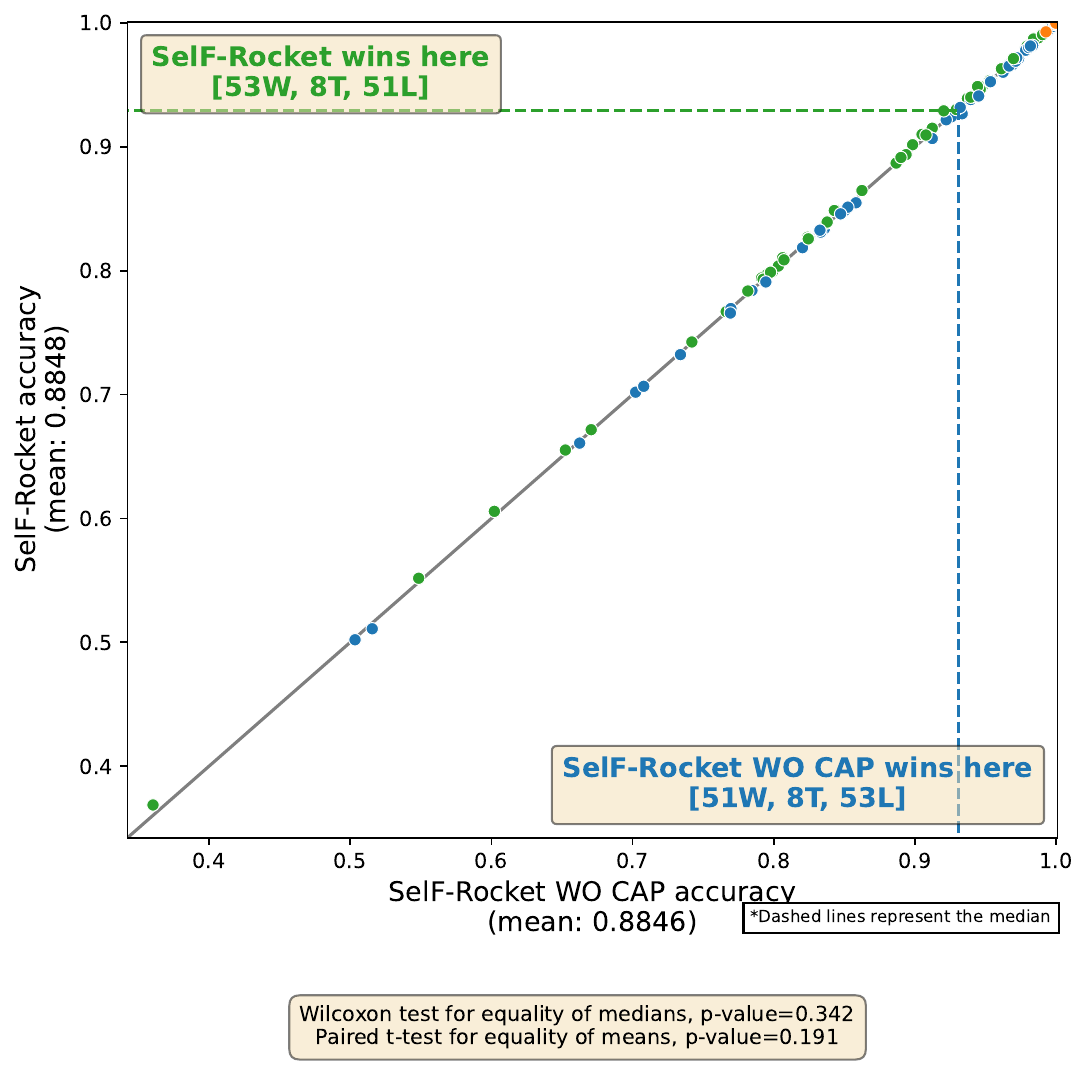}
    \caption{Pairwise accuracy for SelF-Rocket vs SelF-Rocket without a dataset cap in the feature selection module}
    \label{fig:SRWOCAP}
\end{figure}

Table~\ref{tab:DELTACONS} presents the performance differences between SelF-Rocket with and without vote validation, as well as the mean consensus across all resamples for each of the 112 UCR datasets. The mean consensus corresponds to the percentage of voters that agree with the voting decision. A negative $\Delta$ SR/vote indicates that the vote validation system incorrectly overrides the original voting choice. The mean consensus provides an indication of the confidence in the voting outcome: a consensus greater than 90\% (our $vote_{threshold}$) implies that, on average, the original vote is retained, whereas lower values indicate that the vote is, on average, replaced.

\begin{table}[!ht]
   \begin{minipage}{0.6\textwidth}
    \begin{tabular}{ccc} 
    \toprule
        \textbf{Dataset} & \textbf{$\Delta$ SR/vote} & \textbf{Cons} \\ \hline
        \midrule
        ACSF1 & -1.2 & 83.5 \\ \hline
        Adiac & 0.36 & 84 \\ \hline
        ArrowHead & 0.78 & 74 \\ \hline
        Beef & 4.33 & 75.33 \\ \hline
        BeetleFly & 2.17 & 83.67 \\ \hline
        BirdChicken & 0 & 76.33 \\ \hline
        BME & 0 & 92.5 \\ \hline
        Car & -0.5 & 83 \\ \hline
        CBF & -0.06 & 95.5 \\ \hline
        Chinatown & -0.15 & 86.5 \\ \hline
        Chlor.Conc. & -0.37 & 81.67 \\ \hline
        CinCECGTorso & 0.21 & 96 \\ \hline
        Coffee & 0 & 90.5 \\ \hline
        Computers & 0.48 & 91 \\ \hline
        CricketX & -0.18 & 92.33 \\ \hline
        CricketY & -0.2 & 91.83 \\ \hline
        CricketZ & -0.09 & 93 \\ \hline
        Crop & 0.41 & 75.5 \\ \hline
        DiatomSizeRed. & 0.09 & 90.33 \\ \hline
        Dist.Phal.Outl.Age. & 0 & 65.5 \\ \hline
        Dist.Phal.Out.Cor. & 0.5 & 65.33 \\ \hline
        Dist.Phal.TW & 0.34 & 64.17 \\ \hline
        Earthquakes & -0.31 & 66.67 \\ \hline
        ECG200 & 0 & 80.83 \\ \hline
        ECG5000 & 0.1 & 77 \\ \hline
        ECGFiveDays & -0.14 & 83.67 \\ \hline
        ElectricD. & 0.04 & 90.17 \\ \hline
        EOGHoriz. & -0.1 & 94.67 \\ \hline
        EOGVert. & -0.27 & 92.67 \\ \hline
        EthanolLevel & 0 & 99.5 \\ \hline
        FaceAll & 0.02 & 89.5 \\ \hline
        FaceFour & 0.04 & 97.67 \\ \hline
        FacesUCR & -0.11 & 92.67 \\ \hline
        FiftyWords & 0.04 & 97 \\ \hline
        Fish & 0.17 & 77 \\ \hline
        FordA & -0.28 & 95 \\ \hline
        FordB & -0.72 & 81 \\ \hline
        FreezerReg. & 0.01 & 90 \\ \hline
        FreezerSm. & 0.24 & 87 \\ \hline
        GunPoint & 0.29 & 80.67 \\ \hline
        GunPointAg. & 0.09 & 80.83 \\ \hline
        GunPointMal. & 0 & 97.17 \\ \hline
        GunPointOld. & 0 & 100 \\ \hline
        Ham & 1.14 & 76.67 \\ \hline
        HandOutlines & -0.44 & 79.33 \\ \hline
        Haptics & -2.46 & 89.33 \\ \hline
        Herring & -0.68 & 70.67 \\ \hline
        HouseTwenty & 0.62 & 84.33 \\ \hline
        InlineSkate & 0.12 & 77.5 \\ \hline
        InsectEPGReg. & 0 & 98.83 \\ \hline
        InsectEPGSm. & -0.6 & 86.5 \\ \hline
        InsectWing. & 0 & 74.5 \\ \hline
        ItalyPow. & 0.16 & 87.67 \\ \hline
        LargeKit. & 0 & 96 \\ \hline
        Lightning2 & -0.44 & 68.5 \\ \hline
        Lightning7 & 0.87 & 88 \\ \hline
        \vdots & \vdots & \vdots  \\ 
            \bottomrule
            \end{tabular}
            \end{minipage} \hfill
    \begin{minipage}{0.6\textwidth}
    \begin{tabular}{ccc} 
    \toprule
        \textbf{Dataset} & \textbf{$\Delta$ SR/Vote} & \textbf{Cons} \\ \hline
        \midrule
        \vdots & \vdots & \vdots  \\
        Mallat & 0.15 & 92 \\ \hline
        Meat & -0.06 & 97.33 \\ \hline
        MedicalImages & -0.03 & 85.33 \\ \hline
        Mid.Phal.Out.Age. & 0.41 & 55.17 \\ \hline
        Mid.Phal.Out.Cor. & 0.34 & 68.67 \\ \hline
        Mid.Phal.TW & 0.24 & 62.67 \\ \hline
        MixedSh.Reg. & 0.05 & 85.33 \\ \hline
        MixedSh.Sm. & 0.37 & 86 \\ \hline
        MoteStrain & 1.03 & 86 \\ \hline
        NonInv.Fet.ECGT.1 & -0.16 & 90.17 \\ \hline
        NonInv.Fet.ECGT.2 & 0.09 & 81 \\ \hline
        OliveOil & -0.33 & 94.17 \\ \hline
        OSULeaf & -0.01 & 94.33 \\ \hline
        Phal.Out.Cor. & 0.71 & 57.17 \\ \hline
        Phoneme & 1.18 & 90 \\ \hline
        PigAirw. & 0 & 100 \\ \hline
        PigArtP. & 0 & 99.83 \\ \hline
        PigCVP & 0 & 99.67 \\ \hline
        Plane & 0 & 98.17 \\ \hline
        PowerCons & 0.09 & 94 \\ \hline
        Prox.Phal.Out.Age. & 0.05 & 64.17 \\ \hline
        Prox.Phal.Out.Cor. & -0.08 & 71.33 \\ \hline
        Prox.Phal.TW & -0.44 & 66.83 \\ \hline
        Refr.Dev. & 0.44 & 92.5 \\ \hline
        Rock & 1.53 & 81 \\ \hline
        ScreenType & 0 & 97.67 \\ \hline
        SemgHandGen. & 0 & 99.5 \\ \hline
        SemgHandMov. & 0 & 100 \\ \hline
        SemgHandSub. & 0 & 100 \\ \hline
        ShapeletSim & 0 & 96.67 \\ \hline
        ShapesAll & 0 & 98.5 \\ \hline
        SmallKit.Appl. & 0.81 & 78.33 \\ \hline
        SmoothSubspace & 0.04 & 94 \\ \hline
        SonyAIBO.1 & 0.18 & 90.33 \\ \hline
        SonyAIBO.2 & 0.48 & 86.67 \\ \hline
        StarLightCurves & 0.01 & 76.83 \\ \hline
        Strawberry & 0.17 & 73.33 \\ \hline
        SwedishLeaf & 0.11 & 84.67 \\ \hline
        Symbols & 0.38 & 91.33 \\ \hline
        SyntheticControl & 0.04 & 93.33 \\ \hline
        ToeSegmentation1 & 0.19 & 83.33 \\ \hline
        ToeSegmentation2 & 0.49 & 82.33 \\ \hline
        Trace & 0 & 100 \\ \hline
        TwoLeadECG & -0.02 & 90.33 \\ \hline
        TwoPatterns & 0 & 100 \\ \hline
        UMD & 0 & 95.33 \\ \hline
        UWav.All & -0.05 & 95 \\ \hline
        UWaV.X & 0.31 & 81.67 \\ \hline
        UWav.Y & 0.5 & 82.83 \\ \hline
        UWav.Z & 0.15 & 87.33 \\ \hline
        Wafer & 0.05 & 75.67 \\ \hline
        Wine & 1.36 & 81.33 \\ \hline
        WordSynonyms & 0.09 & 91.5 \\ \hline
        Worms & 0.65 & 91.5 \\ \hline
        WormsTwoClass & 1.34 & 75.5 \\ \hline
        Yoga & 0.15 & 87.83 \\ \hline
\bottomrule
\end{tabular}

\end{minipage}
\caption{Difference in accuracy (percentage points) between SelF-ROCKET with and without Vote Validation, and the mean consensus value, averaged over 30 resamples of the 112 UCR datasets.}
\label{tab:DELTACONS}
\end{table}

\end{appendices}

\newpage

\bibliography{bibliography}

\end{document}

%% file: images/Archi.tikz
\begin{circuitikz}
\tikzstyle{every node}=[font=\Large]
\begin{scope}[rotate around={179:(3,13.75)}]
\draw[domain=3:17.25,samples=100,smooth] plot (\x,{2.5*sin(1.91*\x r -1.6 r*rand ) +12.75});
\end{scope}
\draw [line width=1.3pt, ->, >=Stealth] (5.25,15.25) -- (7.75,15.25);
\draw  (10,22) rectangle (11.25,20.75);
\draw  (10,20.75) rectangle (11.25,19.5);
\draw  (10,19.5) rectangle (11.25,18.25);
\draw  (10,18.25) rectangle (11.25,17);
\draw  (10,17) rectangle (11.25,15.75);
\draw  (10,15.75) rectangle (11.25,14.5);
\draw  (10,14.5) rectangle (11.25,13.25);
\draw  (10,13.25) rectangle (11.25,12);
\draw  (10,12) rectangle (11.25,10.75);
\draw  (10,10.75) rectangle (11.25,9.5);
\draw  (10,9.5) rectangle (11.25,8.25);
\draw  (10,8.25) rectangle (11.25,7);
\draw  (11.25,21.75) rectangle (11.5,7);
\draw  (11.5,21.5) rectangle (11.75,7);

\draw  (20,17.75) rectangle (21.25,10.75);
\draw  (21.25,17.5) rectangle (21.75,11.5);
\draw  (21.75,17.25) rectangle (22.25,12.25);
\node [font=\LARGE] at (6.25,14.5) {};
\draw [line width=1.3pt, ->, >=Stealth] (15.25,15.25) -- (17.75,15.25);
\draw [short] (11.75,21.5) -- (20,17.75);
\draw [short] (11.75,7) -- (20,10.75);
\draw [short] (30.25,19.25) -- (30.25,16.75);
\draw [short] (30.25,19.25) -- (32.75,19.25);
\draw [short] (30.75,18.75) -- (30.75,16.25);
\draw [short] (30.75,18.75) -- (33.25,18.75);
\draw [short] (31.25,18.25) -- (31.25,15.75);
\draw [short] (31.25,18.25) -- (33.75,18.25);
\draw [short] (31.75,17.75) -- (31.75,15.25);
\draw [short] (31.75,17.75) -- (34.25,17.75);
\draw [short] (32.25,17.25) -- (32.25,14.75);
\draw [short] (32.25,17.25) -- (34.75,17.25);
\draw [short] (32.75,16.75) -- (32.75,14.25);
\draw [short] (32.75,16.75) -- (35.25,16.75);
\draw [short] (33.25,16.25) -- (33.25,13.75);
\draw [short] (33.25,16.25) -- (35.75,16.25);
\draw [short] (33.75,15.75) -- (33.75,13.25);
\draw [short] (33.75,15.75) -- (36.25,15.75);
\draw [short] (34.25,15.25) -- (34.25,12.75);
\draw [short] (34.25,15.25) -- (36.75,15.25);
\draw [short] (34.75,14.75) -- (34.75,12.25);
\draw [short] (34.75,14.75) -- (37.25,14.75);
\draw [short] (35.25,14.25) -- (35.25,11.75);
\draw [short] (35.25,14.25) -- (37.75,14.25);
\draw [short] (30.25,16.75) -- (30.75,16.75);
\draw [short] (32.75,19.25) -- (32.75,18.75);
\draw [short] (30.75,16.25) -- (31.25,16.25);
\draw [short] (33.25,18.75) -- (33.25,18.25);
\draw [short] (31.25,15.75) -- (31.75,15.75);
\draw [short] (33.75,18.25) -- (33.75,17.75);
\draw [short] (31.75,15.25) -- (32.25,15.25);
\draw [short] (34.25,17.75) -- (34.25,17.25);
\draw [short] (32.25,14.75) -- (32.75,14.75);
\draw [short] (34.75,17.25) -- (34.75,16.75);
\draw [short] (32.75,14.25) -- (33.25,14.25);
\draw [short] (35.25,16.75) -- (35.25,16.25);
\draw [short] (33.25,13.75) -- (33.75,13.75);
\draw [short] (35.75,16.25) -- (35.75,15.75);
\draw [short] (33.75,13.25) -- (34.25,13.25);
\draw [short] (36.25,15.75) -- (36.25,15.25);
\draw [short] (34.25,12.75) -- (34.75,12.75);
\draw [short] (36.75,15.25) -- (36.75,14.75);
\draw [short] (34.75,12.25) -- (35.25,12.25);
\draw [short] (35.25,11.75) -- (37.75,11.75);
\draw [short] (37.75,14.25) -- (37.75,11.75);
\draw [short] (37.25,14.75) -- (37.25,14.25);
\node [font=\LARGE] at (17,15) {};
\node [font=\LARGE] at (17,14.75) {};
\node [font=\LARGE] at (17,15) {};
\node [font=\LARGE] at (17,14.75) {};
\node [font=\LARGE] at (7,14.75) {};
\draw [line width=1.3pt, ->, >=Stealth] (25.75,15.25) -- (28.25,15.25);
\node [font=\LARGE] at (6.75,14) {};
\node [font=\LARGE] at (7.5,14.25) {};
\draw [short] (35.25+10,11.75+2) -- (37.75+10,11.75+2);
\draw [short] (37.75+10,14.25+2) -- (37.75+10,11.75+2);
\draw [short] (35.25+10,11.75+4.5) -- (37.75+10,11.75+4.5);
\draw [short] (37.75+7.5,14.25+2) -- (37.75+7.5,11.75+2);
\draw [line width=1.3pt, ->, >=Stealth] (39,15.25) -- (41.5,15.25);
\node [font=\LARGE] at (6.75,14) {};
\node [font=\LARGE] at (40.75,14.75) {};
\node [font=\LARGE] at (7.25,13.5) {};
\node [font=\LARGE] at (8,13.75) {};
\node [font=\LARGE] at (7.5,14.25) {};
\draw [line width=1.3pt, ->, >=Stealth] (52.5,15.25) -- (55,15.25);
\node [font=\LARGE] at (53.5,17.5) {};
\node [font=\LARGE] at (54.25,17.75) {};
\node [font=\LARGE] at (54,17) {};
\node [font=\LARGE] at (54.75,17.25) {};
\node [font=\LARGE] at (54,17) {};
\node [font=\LARGE] at (54.5,16.5) {};
\node [font=\LARGE] at (55.25,16.75) {};
\node [font=\LARGE] at (54.75,17.25) {};
\draw [rounded corners = 9.0] (58.75,20.5) rectangle  node {\Huge Ridge} (63.5,11.25);

\node[scale=2.0] [font={\Huge\bfseries\sffamily}] at (-4.75,23) {\textit{\textbf{Time series}}};
\node[scale=2.0] [font={\Huge\bfseries\sffamily}] at (20.25,25) {\textit{\textbf{Random 1-D}}};
\node[scale=2.0] [font={\Huge\bfseries\sffamily}] at (20.25,23) {\textit{\textbf{Convolution kernels}}};
\node[scale=2.0] [font={\Huge\bfseries\sffamily}] at (10.25,5) {\textit{\textbf{Input}}};
\node[scale=2.0] [font={\Huge\bfseries\sffamily}] at (10.25,3) {\textit{\textbf{Representation}}};
\node[scale=2.0] [font={\Huge\bfseries\sffamily}] at (35.75,5) {\textit{\textbf{Pooling Operators}}};
\node[scale=2.0] [font={\Huge\bfseries\sffamily}] at (35.75,3) {\textit{\textbf{(PPV, ZC, ...)}}};
\node[scale=2.0] [font={\Huge\bfseries\sffamily}] at (45.75,25) {\textit{\textbf{Feature}}};
\node[scale=2.0] [font={\Huge\bfseries\sffamily}] at (45.75,23) {\textit{\textbf{Selection}}};
\node[scale=2.0] [font={\Huge\bfseries\sffamily}] at (61,8) {\textit{\textbf{Training}}};
\end{circuitikz}

%% file: images/SROV.tikz
\tikzset{
  dataset/.pic = {
    \path (0,0) coordinate (O);
    \begin{scope}[transform canvas = {scale around={#1:(O)}}]
        \draw[line width=3mm] (0,0) ellipse (2 and .6);
        \draw[line width=3mm] (-2,0) -- (-2,-3);
        \draw[line width=3mm] (2,0) -- (2,-3);
        \draw[line width=2mm] (-2,-1) arc (180:360:2 and 0.6);
        \draw[line width=2mm] (-2,-2) arc (180:360:2 and 0.6);
        \draw[line width=3mm] (-2,-3) arc (180:360:2 and 0.6);
        \filldraw[black] (1,-1) circle (0.2);
        \filldraw[black] (1,-2) circle (0.2);
        \filldraw[black] (1,-3) circle (0.2);
    \end{scope}
  }, 
  pics/dataset/.default = 1,
  ml/.pic = {
    \path (0,0) coordinate (O);
    \begin{scope}[transform canvas = {scale around={#1:(O)}}]
        \draw[-{Triangle[length=5mm,width=6mm]}, line cap=round, line width=2mm] (0,-.3) --++ (90:3.2);
        \draw[-{Triangle[length=5mm,width=6mm]}, line cap=round, line width=2mm] (-.3,0) --++ (0:3.2);
        \draw[line cap=round, line width=2mm] (0,0) --++ (45:3.2);
        \draw[fill=white, line width=1.8mm] (.8,1.8) circle (0.32);
        \draw[fill=white, line width=1.8mm] (1,.3) --++(0:.6) --++(120:.6) -- cycle;
        \draw[fill=white, line width=2mm] (1.8,1) --++(0:.6) --++(120:.6) -- cycle;
    \end{scope}
  }, 
  pics/ml/.default = 1,
  mlc/.pic = {
    \path (0,0) coordinate (O);
    \begin{scope}[transform canvas = {scale around={#1:(O)}}]
        \draw[-{Triangle[length=5mm,width=6mm]}, line cap=round, line width=2mm] (0,-.3) --++ (90:3.2);
        \draw[-{Triangle[length=5mm,width=6mm]}, line cap=round, line width=2mm] (-.3,0) --++ (0:3.2);
        \draw[line cap=round, line width=2mm] (0,0) --++ (45:3.2);
        \draw[fill=white, line width=1.8mm] (.8,1.8) circle (0.32);
        \draw[fill=white, line width=1.8mm] (1,.3) --++(0:.6) --++(120:.6) -- cycle;
        \draw[fill=white, line width=2mm] (1.8,1) --++(0:.6) --++(120:.6) -- cycle;
        \draw[line width=1mm] (1.2,1.2) circle (2.2cm);
    \end{scope}
  },
  pics/mlc/.default = 1,
  mllink/.pic = {
    \path (0,0) coordinate (O);
    \begin{scope}[transform canvas = {scale around={#1:(O)}}]
        \draw[line cap=round, >={Stealth[length=18pt]}, line width=1.8mm, ->, rounded corners=10pt] (0,0) -- ++(0:6);
        \draw[fill=white, line width=1.5mm] ($(0,0)!0.5!(6,0)$) circle (0.8);
        \pic at (2.65, -.35) {ml=.3};
    \end{scope}
  }, 
  pics/mllink/.default = 1,
  split/.pic = {
    \path (0,0) coordinate (O);
    \begin{scope}[transform canvas = {scale around={#1:(O)}}]
        \draw[rounded corners=15pt, line width=1mm] (-1.2,0) rectangle (1.2,4); 
        \pic at (0,1.7) {dataset=.3};
        \node at (0,2.2) {TRAIN};
        \pic at (0,3.6) {dataset=.3};
        \node at (0,.3) {VALIDAT.};
    \end{scope}
  }, 
  pics/split/.default = 1,
}
    
  \begin{tikzpicture}
    \tikzmath{
        \x0 = 0;       
        \x1 = 4.8;     
        \x2 = 6.2;     
        \x3 = \x2+1.5; 
        \x4 = 11.8;    
        \x5 = 18.7;    
        \x6 = 25.4;    
        \y0 = 13.5;    
        \y1 = 12.5;    
        \y2 = 11.85;   
        \y3 = 11;      
        \y4 = 8.7;     
        \y5 = 7.7;     
        \y6 = 7.05;    
        \y7 = 6.2;     
        \y8 = 3.1;     
        \y9 = 2.1;     
        \ya = 1.45;   
        \yb = .6;     
    }
    
    \node[text width = 4cm, text centered] at (\x0, 9.1) {
        ORIGINAL \\
        TRAIN \\
        DATASET};
    \pic at (\x0, 8) {dataset=.5};

    \node[text width = 4cm, text centered] at (\x1, 15.6) {};
    \node at (\x1, 14.7) {SPLIT (VOTER) \#$1$};
    \pic at (\x1, 10.4) {split=1};
    \node at (\x1, 9.9) {SPLIT (VOTER) \#$2$};
    \pic at (\x1, 5.6) {split=1};
    \node at (\x1, 5.1) {$\bm{\vdots}$};
    \node at (\x1, 4.3) {SPLIT (VOTER) \#$\bm{k \times nr}$};
    \pic at (\x1, 0) {split=1};

    \draw[line cap=round, >={Stealth[length=10pt]}, line width=.8mm, ->, rounded corners=10pt] (1.4, 7) --++(0:1) --++(270:4) --++(0:1);
    \draw[line cap=round, >={Stealth[length=10pt]}, line width=.8mm, ->, rounded corners=10pt] (1.4, 7) --++(0:1) --++(90:4) --++(0:1);
    \draw[line cap=round, >={Stealth[length=10pt]}, line width=.8mm, ->, rounded corners=10pt] (1.4, 7) --++(0:1) --++(90:1) --++(0:1);
    
    \pic at (\x2, \y0) {mllink=.5};
    \pic at (\x2, \y1) {mllink=.5};
    \node at (\x3, \y2) {$\bm{\vdots}$};
    \pic at (\x2, \y3) {mllink=.5};
    \pic at (\x2, \y4) {mllink=.5};
    \pic at (\x2, \y5) {mllink=.5};
    \node at (\x3, \y6) {$\bm{\vdots}$};
    \pic at (\x2, \y7) {mllink=.5};
    \pic at (\x2, \y8) {mllink=.5};
    \pic at (\x2, \y9) {mllink=.5};
    \node at (\x3, \ya) {$\bm{\vdots}$};
    \pic at (\x2, \yb) {mllink=.5};

    \node[text width = 4cm, text centered] at (\x4, 14.8) {
        PERFORMANCE \\
        MEASUREMENT};
    \node[draw, line width=.6mm, minimum width=4.8cm, minimum height=.6cm, align=center, rounded corners=5pt]  at (\x4, \y0) {
        ACCURACY IR-PO \#$1$
    };
    \node[draw, line width=.6mm, minimum width=4.8cm, minimum height=.6cm, align=center, rounded corners=5pt]  at (\x4, \y1) {
        ACCURACY IR-PO \#$2$
    };
    \node at (\x4, \y2) {$\bm{\vdots}$};
    \node[draw, line width=.6mm, minimum width=4.8cm, minimum height=.6cm, align=center, rounded corners=5pt]  at (\x4, \y3) {
        ACCURACY IR-PO \#$N$
    };
    \node[draw, line width=.6mm, minimum width=4.8cm, minimum height=.6cm, align=center, rounded corners=5pt]  at (\x4, \y4) {
        ACCURACY IR-PO \#$1$
    };
    \node[draw, line width=.6mm, minimum width=4.8cm, minimum height=.6cm, align=center, rounded corners=5pt]  at (\x4, \y5) {
        ACCURACY IR-PO \#$2$
    };
    \node at (\x4, \y6) {$\bm{\vdots}$};
    \node[draw, line width=.6mm, minimum width=4.8cm, minimum height=.6cm, align=center, rounded corners=5pt]  at (\x4, \y7) {
        ACCURACY IR-PO \#$N$
    };
    \node at (\x4, 5.1) {$\bm{\vdots}$};
    \node[draw, line width=.6mm, minimum width=4.8cm, minimum height=.6cm, align=center, rounded corners=5pt]  at (\x4, \y8) {
        ACCURACY IR-PO \#$1$
    };
    \node[draw, line width=.6mm, minimum width=4.8cm, minimum height=.6cm, align=center, rounded corners=5pt]  at (\x4, \y9) {
        ACCURACY IR-PO \#$2$
    };
    \node at (\x4, \ya) {$\bm{\vdots}$};
    \node[draw, line width=.6mm, minimum width=4.8cm, minimum height=.6cm, align=center, rounded corners=5pt]  at (\x4, \yb) {
        ACCURACY IR-PO \#$N$
    };

    \node[text width = 4cm, text centered] at (\x5, {\y0+.3}) {
        IR-PO \#$1$};
    \node[draw, line width=.6mm, minimum width=4.8cm, minimum height=.6cm, align=center, rounded corners=5pt]  at (\x5, \y1) {
        ACCURACY SPLIT \#$1$ \\
        ACCURACY SPLIT \#$2$ \\
        $\bm{\vdots}$ \\
        ACCURACY SPLIT \#$k \times nr$
    };
    \node[text width = 4cm, text centered] at (\x5, {\y4+.3}) {
        IR-PO \#$2$};
    \node[draw, line width=.6mm, minimum width=4.8cm, minimum height=.6cm, align=center, rounded corners=5pt]  at (\x5, \y5) {
        ACCURACY SPLIT \#$1$ \\
        ACCURACY SPLIT \#$2$ \\
        $\bm{\vdots}$ \\
        ACCURACY SPLIT \#$k \times nr$
    };
    \node at (\x5, 5.1) {$\bm{\vdots}$};
    \node[text width = 4cm, text centered] at (\x5, {\y8+.3}) {
        IR-PO \#$N$};
    \node[draw, line width=.6mm, minimum width=4.8cm, minimum height=.6cm, align=center, rounded corners=5pt]  at (\x5, \y9) {
        ACCURACY SPLIT \#$1$ \\
        ACCURACY SPLIT \#$2$ \\
        $\bm{\vdots}$ \\
        ACCURACY SPLIT \#$k \times nr$
    };

    \draw[line cap=round, >={Stealth[length=10pt]}, line width=.8mm, ->, rounded corners=10pt] ({\x4+2.6}, \y0) --++(0:1) --++(270:1) --++(0:.75);
    \draw[line cap=round, >={Stealth[length=10pt]}, line width=.8mm, ->, rounded corners=10pt] ({\x4+2.6}, \y1) --++(0:.6) --++(270:4.8) --++(0:1.15);
    \draw[line cap=round, >={Stealth[length=10pt]}, line width=.8mm, ->, rounded corners=10pt] ({\x4+2.6}, \y3) --++(0:.25) --++(270:8.9) --++(0:1.55);
    \draw[line cap=round, >={Stealth[length=10pt]}, line width=.8mm, ->, rounded corners=10pt] ({\x4+2.6}, \y4) --++(0:1) --++(90:3.8) --++(0:.75);
    \draw[line cap=round, >={Stealth[length=10pt]}, line width=.8mm, ->, rounded corners=10pt] ({\x4+2.6}, \y5) --++(0:1.75);
    \draw[line cap=round, >={Stealth[length=10pt]}, line width=.8mm, ->, rounded corners=10pt] ({\x4+2.6}, \y7) --++(0:.25) --++(270:4.1) --++(0:1.55);
    \draw[line cap=round, >={Stealth[length=10pt]}, line width=.8mm, ->, rounded corners=10pt] ({\x4+2.6}, \y8) --++(0:1) --++(90:9.4) --++(0:.75);
    \draw[line cap=round, >={Stealth[length=10pt]}, line width=.8mm, ->, rounded corners=10pt] ({\x4+2.6}, \y9) --++(0:.6) --++(90:5.6) --++(0:1.15);
    \draw[line cap=round, >={Stealth[length=10pt]}, line width=.8mm, ->, rounded corners=10pt] ({\x4+2.6}, \yb) --++(0:.25) --++(90:1.49) --++(0:1.55);

    \node[draw, line width=.6mm, minimum width=4.8cm, minimum height=.6cm, align=center, rounded corners=5pt]  at (\x6, {\y5-.6}) {
        HIGHEST MEDIAN \\
        VOTE \\
        and \\
        VOTE VALIDATION
    };

    \draw[line cap=round, >={Stealth[length=10pt]}, line width=.8mm, ->, rounded corners=10pt] (21.35, \y1) --++(0:.75) --++(270:5.5) --++(0:.75);
    \draw[line cap=round, >={Stealth[length=10pt]}, line width=.8mm, ->, rounded corners=10pt] (21.35, \y5) --++(0:.75) --++(270:.7) --++(0:.75);
    \draw[line cap=round, >={Stealth[length=10pt]}, line width=.8mm, ->, rounded corners=10pt] (21.35, \y9) --++(0:.75) --++(90:4.9) --++(0:.75);
    
  \end{tikzpicture}